%% file: main.tex
\documentclass{article}

\input{preamble_pkgs}
\input{preamble_macros}

\usepackage{graphbox} % additional options for includegraphics, e.g. align

\newcommand{\eg}{\textit{e.\,g., }}

\newcommand{\ie}{\textit{i.\,e., }}

\mathchardef\mhyphen="2D
\usepackage{layouts}

\usepackage[accepted]{arxiv}

\icmltitlerunning{Geometric Dataset Distances via Optimal Transport}

\begin{document}

\twocolumn[
\icmltitle{Geometric Dataset Distances via Optimal Transport}




\begin{icmlauthorlist}
\icmlauthor{David Alvarez-Melis}{msr}
\icmlauthor{Nicol\`o Fusi}{msr}
\end{icmlauthorlist}

\icmlaffiliation{msr}{Microsoft Research}

\icmlcorrespondingauthor{David Alvarez-Melis}{alvarez.melis@microsoft.com}
\icmlcorrespondingauthor{Nicol\`o Fusi}{fusi@microsoft.com}

\icmlkeywords{Optimal Transport, Machine Learning, Dataset Distance}

\vskip 0.3in
]



\printAffiliationsAndNotice{}  % leave blank if no need to mention equal contribution

\input{intro}       % Abstract + Intro
\input{related}     % Related work
\input{background}  % Background 
\input{approach}    % Main section
\input{computation} % Computational Analysis and Optimization 
\input{experiments} % All experiments
\input{conclusion}  % 

\pagebreak
\clearpage

\bibliography{references.bib}
\bibliographystyle{icml2020}

\raggedbottom

\pagebreak
\clearpage

\input{appendix}

\end{document}

%% file: preamble_pkgs.tex
\usepackage{etoolbox}
\providebool{loadgeometry} % will default to false
\providebool{altermargins} % will default to false
\providebool{usebiblatex}  % will default to false

\setbool{loadgeometry}{false}
\setbool{usebiblatex}{false}

\usepackage[T1]{fontenc}
\usepackage{tgtermes}

\usepackage{amssymb}

\usepackage{scalefnt,letltxmacro}
\LetLtxMacro{\oldtextsc}{\textsc}
\renewcommand{\textsc}[1]{\oldtextsc{\scalefont{1.10}#1}}
\usepackage[ttdefault=true]{AnonymousPro}

\usepackage{amsmath}
\usepackage{amsmath,amsthm, mathtools}
\usepackage{amscd}
\usepackage{bbm}
\usepackage{commath}      % Differentials, etc
\usepackage{dsfont}       % For \mathds
\usepackage{upgreek}      % For \upalpha etc
\usepackage{stmaryrd}     % For nice double bracket [[ ]] 
\usepackage{siunitx}      % Scientific notation
\usepackage{xspace}
\usepackage{float}        % For H in tables/figs

\ifbool{loadgeometry}{
	\ifbool{altermargins}{
		\usepackage[
		  paper  = letterpaper,
		  left   = 1.25in,
		  right  = 1.25in,
		  top    = 1.0in,
		  bottom = 1.0in,
		  ]{geometry}
	}{
		\usepackage{geometry} 	% For conference/journal papers, don't mess with margins
	}
}

\usepackage[english]{babel}
\usepackage[parfill]{parskip}
\usepackage{afterpage}
\usepackage{enumitem}
\usepackage{framed}
\usepackage{xspace}
\usepackage{xparse}% http://ctan.org/pkg/xparse

\usepackage{lineno}

\usepackage{ragged2e}

\newcounter{parcount}

\ifbool{altermargins}{
	\setlength{\marginparwidth}{1.25in}
}{}

\usepackage[usenames,dvipsnames]{xcolor}
\definecolor{shadecolor}{gray}{0.9}

\usepackage[pdftex]{graphicx}
\usepackage[labelfont=bf]{caption}
\usepackage[format=hang]{subcaption}
\usepackage{booktabs, array}

\usepackage[algoruled]{algorithm2e}
\usepackage{listings}
\usepackage{fancyvrb}
\fvset{fontsize=\small}

\usepackage[colorlinks,linktoc=all]{hyperref}
\hypersetup{citecolor=MidnightBlue}
\hypersetup{linkcolor=MidnightBlue}
\hypersetup{urlcolor=MidnightBlue}
\usepackage[nameinlink]{cleveref}
\creflabelformat{equation}{#2\textup{#1}#3}  % <- remove parenthesis from eqs
\crefname{section}{§}{§§}
\Crefname{section}{§}{§§}

\usepackage
[acronym,smallcaps,nowarn,section,nogroupskip,nonumberlist]{glossaries}
\glsdisablehyper{}

\lstdefinestyle{mystyle}{
    commentstyle=\color{OliveGreen},
    numberstyle=\tiny\color{black!60},
    stringstyle=\color{BrickRed},
    basicstyle=\ttfamily\scriptsize,
    breakatwhitespace=false,
    breaklines=true,
    captionpos=b,
    keepspaces=true,
    numbers=none,
    numbersep=5pt,
    showspaces=false,
    showstringspaces=false,
    showtabs=false,
    tabsize=2
}
\ifbool{usebiblatex}{
	\usepackage[
	    backend=biber,
		citestyle=authoryear,
	    defernumbers=true,       % continuous numbering across bibliography sections
		firstinits=true,         % abbreviate first name
		doi=false,
		natbib=true,
		isbn=false,
		eprint=false,
		url=false,
	]{biblatex}
}{
	\usepackage{natbib}
}

\usepackage{xargs}                      % Use more than one optional parameter in a new commands
\usepackage[colorinlistoftodos,prependcaption,textsize=tiny]{todonotes}
\newcommandx{\unsure}[2][1=]{\todo[linecolor=red,backgroundcolor=red!25,bordercolor=red,#1]{#2}}
\newcommandx{\change}[2][1=]{\todo[linecolor=blue,backgroundcolor=blue!25,bordercolor=blue,#1]{#2}}
\newcommandx{\info}[2][1=]{\todo[linecolor=OliveGreen,backgroundcolor=OliveGreen!25,bordercolor=OliveGreen,#1]{#2}}
\newcommandx{\improvement}[2][1=]{\todo[linecolor=Plum,backgroundcolor=Plum!25,bordercolor=Plum,#1]{#2}}
\newcommandx{\thiswillnotshow}[2][1=]{\todo[disable,#1]{#2}}

%% file: preamble_macros.tex
\newtheorem{theorem}{Theorem}[section]

\newtheorem{proposition}[theorem]{Proposition}

\newtheorem{lemma}[theorem]{Lemma}
\newtheorem{lemma*}{Lemma}

\def\R{\mathbb{R}}
\def\C{\mathbb{C}}
\def\K{\mathbb{K}}

\def\K{\mathbf{K}}

\def\u{\mathbf{u}}
\def\v{\mathbf{v}}

\def\x{\mathbf{x}}
\def\y{\mathbf{y}}

\def\0{\mathbf{0}}

\def\cD{\mathcal{D}}

\def\cN{\mathcal{N}}

\def\cP{\mathcal{P}}

\def\cT{\mathcal{T}}

\def\cX{\mathcal{X}}
\def\cY{\mathcal{Y}}
\def\cZ{\mathcal{Z}}

\DeclareMathOperator{\tr}{tr}
\DeclarePairedDelimiterX{\klx}[2]{(}{)}{#1\;\delimsize\|\;#2}

\DeclarePairedDelimiterX{\relentx}[2]{(}{)}{#1\;\delimsize|\;#2}

\DeclarePairedDelimiterX{\inner}[2]{\langle}{\rangle}{#1, #2} % inner prod

\newcommand{\suchthat}{\;\ifnum\currentgrouptype=16 \middle\fi|\;}

\def\OT{\textup{OT}}

\def\W{\textup{W}}                      % Wasserstein Distance
\makeatletter
\renewcommand*\env@matrix[1][*\c@MaxMatrixCols c]{%
  \hskip -\arraycolsep
  \let\@ifnextchar\new@ifnextchar
  \array{#1}}
\makeatother

\newenvironment{itemize*}%
  {\begin{itemize}%
  \vspace{-0.5cm}
    \setlength{\itemsep}{0pt}%
    \setlength{\parskip}{0pt}}%
  {\end{itemize}}
\newenvironment{enumerate*}%
{\begin{enumerate}
    \vspace{-0.5cm}
    \setlength{\itemsep}{0pt}%
    \setlength{\parskip}{0pt}}%
  {\end{enumerate}}

\newcommand{\mnist}{\textsc{mnist}\xspace}

\newcommand{\usps}{\textsc{usps}\xspace}

%% file: intro.tex
\begin{abstract}
The notion of task similarity is at the core of various machine learning paradigms, such as domain adaptation and meta-learning. Current methods to quantify it are often heuristic, make strong assumptions on the label sets across the tasks, and many are architecture-dependent, relying on task-specific optimal parameters (\eg require training a model on each dataset). In this work we propose an alternative notion of distance between datasets that (i) is model-agnostic, (ii) does not involve training, (iii) can compare datasets even if their label sets are completely disjoint and (iv) has solid theoretical footing. This distance relies on optimal transport, which provides it with rich geometry awareness, interpretable correspondences and well-understood properties. Our results show that this novel distance provides meaningful comparison of datasets, and correlates well with transfer learning hardness across various experimental settings and datasets. 
\end{abstract}

\section{Introduction}

A key hallmark of machine learning practice is that labeled data from the application of interest is usually scarce. For this reason, there is vast interest in methods that can combine, adapt and transfer knowledge across datasets and domains. Entire research areas are devoted to these goals, such as domain adaptation, transfer-learning and meta-learning. A fundamental concept underlying all these paradigms is the notion of \textit{distance} (or more generally, \textit{similarity}) between datasets. For instance, transferring knowledge across similar domains should intuitively be easier than across distant ones. Likewise, given a choice of various datasets to pretrain a model on, it would seem natural to choose the one that is closest to the task of interest.

Despite its evident usefulness and apparent simpleness, the notion of distance between datasets is an elusive one, and quantifying it efficiently and in a principled manner remains largely an open problem. Doing so requires solving various challenges that commonly arise precisely in the settings for which this notion would be most useful, such as the ones mentioned above. For example, in supervised machine learning settings the datasets consist of both features and labels, and while defining a distance between the former is often\;---though not always---\;trivial, doing so for the labels is far from it, particularly if the label-sets across the two tasks are not identical (as is often the case for off-the-shelf pretrained models). 

Current approaches to transfer learning that seek to quantify dataset similarity circumvent these challenges in various ingenious, albeit often heuristic, ways. A common approach is to compare the dataset via proxies, such as the learning curves of a pre-specified model \citep{leite2005predicting} or its optimal parameters \citep{achille2019task, khodak2019adaptive} on a given task, or by making strong assumptions on the similarity or co-occurrence of labels across the two datasets \citep{tran2019transferability}. Most of these approaches lack guarantees, are highly dependent on the probe model used, and require training a model to completion (\eg to find optimal parameters) on each dataset being compared. On the opposite side of the spectrum are principled notions of discrepancy between domains \cite{ben-david2007analysis, mansour2009domain}, which nevertheless are often not computable in practice, or do not scale to the type of datasets used in machine learning practice. 

In this work, we seek to address some of these limitations by proposing an alternative notion of distance between datasets. At the heart of this approach is the use of optimal transport (OT) distances \citep{villani2008optimal} to compare distributions over feature-label pairs in a geometrically-meaningful and principled way. In particular, we propose a hybrid Euclidean-Wasserstein distance between feature-label pairs across domains, where labels themselves are modeled as distributions over features vectors. As a consequence of this technique, our framework allows for comparison of datasets \textit{even if their label sets are completely unrelated or disjoint}, as long as a distance between their features can be defined. This notion of distance between labels, a by-product of our approach, has itself various potential uses, \eg to optimally sub-sample classes from large datasets for more efficient pretraining. 

In summary, we make the following contributions:
\vspace{-0.2cm}
\begin{itemize}[wide=0pt, leftmargin=\dimexpr\labelwidth + 2\labelsep\relax]
    \setlength\itemsep{-0.5em}
	\item We introduce a notion of distance between datasets that is principled, flexible and computable in practice
    \item We propose various algorithmic shortcuts to scale up computation of this distance to very large datasets
    \item We provide extensive empirical evidence that this distance is highly predictive of transfer learning success across various domains, tasks and data modalities
\end{itemize}

%% file: related.tex
%!TEX root = main.tex

\section{Related Work}

\paragraph{Discrepancy Distance} Various notions of (dis)similarity between data distributions have been proposed in the context of domain adaptation, such as the $d_A$ \citep{ben-david2007analysis} and discrepancy distances\footnote{Despite its name, this discrepancy is not a distance in general.} \citep{mansour2009domain}. These discrepancies depend on a loss function and hypothesis (\ie predictor) class, and quantify dissimilarity through a supremum over this function class. The latter discrepancy in particular has proven remarkably useful for proving generalization bounds for adaptation \citep{cortes2011domain}, and while it can be estimated from samples, bounding the approximation quality relies on quantities like the VC-dimension of the hypothesis class, which might not be always known or easy to compute. 

\vspace{-0.2cm}
\paragraph{Dataset Distance via Parameter Sensitivity}
The Fisher information metric is a classic notion from information geometry \citep{amari1985differential, amari2000methods} that characterizes a parametrized probability distribution locally through the sensitivity of its density to changes in the parameters. In  machine learning, it has been used to analyze and improve optimization approaches \citep{amari1998natural} and to measure the capacity of neural networks \citep{liang2019fisher-rao}. In recent work, \citet{achille2019task} use this notion to construct vector representations of tasks, which they then use to define a notion of similarity between these. They show that this notion recovers taxonomic similarities and is useful in meta-learning to predict whether a certain feature extractor will perform well in a new task. While this notion shares with ours its agnosticism of the number of classes and their semantics, it differs in the fact that it relies on a probe network trained on a specific dataset, so its geometry is heavily influenced by the characteristics of this network. Besides the Fisher information, a related information-theoretic notion of complexity that can be used to characterize tasks is the Kolmogorov Structure Function \citep{li2006data}, which \citet{achille2018dynamics} use to define a notion of \textit{reachability} between tasks. 

\vspace{-0.2cm}
\paragraph{Optimal Transport-based distributional distances}
The general idea of representing complex objects via distributions, which are then compared through optimal transport distances, is an active area of research. Also driven by the appeal of their closed-form Wasserstein distance, \citet{muzellec2018generalizing} propose to embed objects as elliptical distributions, which requires differentiating through these distances, and discuss various approximations to scale up these computations. \citet{frogner2019learning} extend this idea but represent the embeddings as discrete measures (\ie point clouds) rather than Gaussian/Elliptical distributions. Both of these works focus on embedding and consider only within-dataset comparisons. Also within this line of work, \citet{delon2019wasserstein} introduce a Wasserstein-type distance between Gaussian mixture models. Their approach restricts the admissible transportation couplings themselves to be Gaussian mixture models, and does not directly model label-to-label similarity. More generally, the Gromov-Wasserstein distance \citep{memoli2011gromov} has been proposed to compare collections across different domains \citep{memoli2017distances, alvarez-melis2018gromov}, albeit leveraging only features, not labels.

\vspace{-0.2cm}
\paragraph{Hierarchical OT distances} 
The distance we propose can be understood as a hierarchical OT distance, \ie one where the ground metric itself is defined through an OT problem. This principle has been explored in other contexts before. For example, \citet{yurochkin2019hierarchical} use a hierarchical OT distance for document similarity, defining a inner-level distance between topics and a outer-level distance between documents using OT. \citep{dukler2019wasserstein} on the other hand use a nested Wasserstein distance as a loss for generative model training, motivated by the observation that the Wasserstein distance is better suited to comparing images than the usual pixel-wise $L_2$ metric used as ground metric. Both the goal, and the actual metric, used by these approaches differs from ours.

\vspace{-0.2cm}
\paragraph{Optimal Transport for Domain Adaptation}
Using label information to guide the optimal transport problem towards class-coherent matches has been explored before, \eg by enforcing group-norm penalties \citep{courty2017optimal} or through submodular cost functions \citep{alvarez-melis2018structured}. These works are focused on the unsupervised domain adaptation setting, so their proposed modifications to the OT objective use only label information from one of the two domains, and even then, do so without explicitly defining a metric between these. Furthermore, they do not lead to proper distances, and these works deal with a single static pair of tasks, so they lack analysis of the distance across multiple source and target datasets. 

%% file: background.tex
\section{Background on Optimal Transport}\label{sec:background}

Optimal transport (OT) is a powerful and principled approach to compare probability distributions, with deep roots in statistics, computer science and applied mathematics \citep{villani2003topics, villani2008optimal}. Among many desirable properties, these distances leverage the geometry of the underlying space, making them ideal for comparing distributions, shapes and point clouds \citep{peyre2019computational}. 

The OT problem considers a complete and separable metric space $\cX$, along with probability measures $\alpha \in \cP(\cX)$ and $\beta \in \cP(\cX)$. These can be continuous or discrete measures, the latter often used in practice as empirical approximations of the former whenever working in the finite-sample regime. The Kantorovich formulation \cite{kantorovitch1942translocation} of the transportation problem reads:
\begin{equation}\label{eq:ot_problem}
    \text{OT}(\alpha, \beta) \triangleq \min_{\pi \in \Pi(\alpha, \beta)} \int_{\cX \times \cX} c(x,y) \dif \pi(x,y),
\end{equation}
where $c(\cdot, \cdot): \cX \times \cX \rightarrow \R^{+}$ is a cost function (the ``ground'' cost), and the set of couplings $\Pi(\alpha, \beta)$ consists of joint probability distributions over the product space $\cX \times \cX$ with marginals $\alpha$ and $\beta$, that is,
\begin{equation}\label{eq:transportation_polytope}
    \Pi(\alpha, \beta) \triangleq \{\pi \in \cP(\cX\!\times\!\cX) \suchthat P_{1\#}\pi = \alpha,  P_{2\#}\pi=\beta \}.
\end{equation}
Whenever $\cX$ is equipped with a metric $d_{\cX}$, it is natural to use it as ground cost, \eg $c(x,y) = d_{\cX}(x,y)^p$ for some $p \geq 1$. In such case, $\text{W}_p(\alpha,\beta)\triangleq \text{OT}(\alpha,\beta)^{1/p}$ is called the $p$-Wasserstein distance. The case $p=1$ is also known as the Earth Mover's Distance \citep{rubner2000earth}.

The measures $\alpha$ and $\beta$ are rarely known in practice. Instead, one has access to finite samples $\{\x^{(i)}\} \in \cX, \{\y^{(j)}\} \in \cX$. In that case, one can construct discrete measures $\alpha=\sum_{i=1}^n \mathbf{a}_i\delta_{\x^{(i)}}$ and $\beta=\sum_{i=1}^m \mathbf{b}_i\delta_{\y^{(j)}}$, where $\mathbf{a}$, $\mathbf{b}$ are vectors in the probability simplex, and the pairwise costs can be compactly represented as an $n\times m$ matrix $\mathbf{C}$, \ie $\mathbf{C}_{ij}=c(\x^{(i)}, \y^{(j)})$. In this case, Equation \eqref{eq:ot_problem} becomes a linear program. Solving this problem scales cubically on the sample sizes, which is often prohibitive in practice. Adding an entropy regularization, namely
\begin{equation}\label{eq:entreg_wasserstein}
       \! \text{OT}_{\epsilon}(\alpha, \beta) \triangleq \hspace{-0.85em}\min_{\pi \in \Pi(\alpha, \beta)}\! \int_{\cX^2} \hspace{-0.75em} c(x,y) \dif \pi(x,y) + \epsilon \text{H}(\pi\!\mid\!\alpha \otimes \beta),
\end{equation}
where $\text{H}(\pi\mid \alpha\otimes \beta)=\int \log(\dif \pi / \dif \alpha \dif \beta)\dif \pi $ is the relative entropy, leads to a problem that can be solved much more efficiently \citep{cuturi2013sinkhorn,altschuler2017nearlinear} and with better sample complexity \citep{genevay2019sample} than the original one. The \textit{Sinkhorn divergence} \citep{genevay2018learning}, defined as 
\begin{equation*}
    \textup{SD}_{\varepsilon}(\alpha, \beta) = \OT_{\varepsilon}(\alpha, \beta) -\frac{1}{2}\OT_{\varepsilon}(\alpha, \alpha) - \frac{1}{2}\OT_{\varepsilon}(\beta, \beta),
\end{equation*}
has various desirable properties, \eg it is positive, convex and metrizes the weak$^{\ast}$ convergence of distributions \citep{feydy2019interpolating}.

In the discrete case, problem~\eqref{eq:entreg_wasserstein} can be solved with the Sinkhorn algorithm \citep{cuturi2013sinkhorn,peyre2019computational}, a matrix-scaling procedure which iteratively updates $\u \gets \mathbf{a} \oslash \K \v$ and $\v \gets \mathbf{b} \oslash \K^{\top}\u$, where $\K\triangleq \exp\{-\frac{1}{\epsilon}\C\}$ and the division $\oslash$ and exponential are entry-wise. 

\begin{figure}
    \centering
    \includegraphics[width=0.8\linewidth, trim = {1cm 1cm 1cm 1cm}, clip]{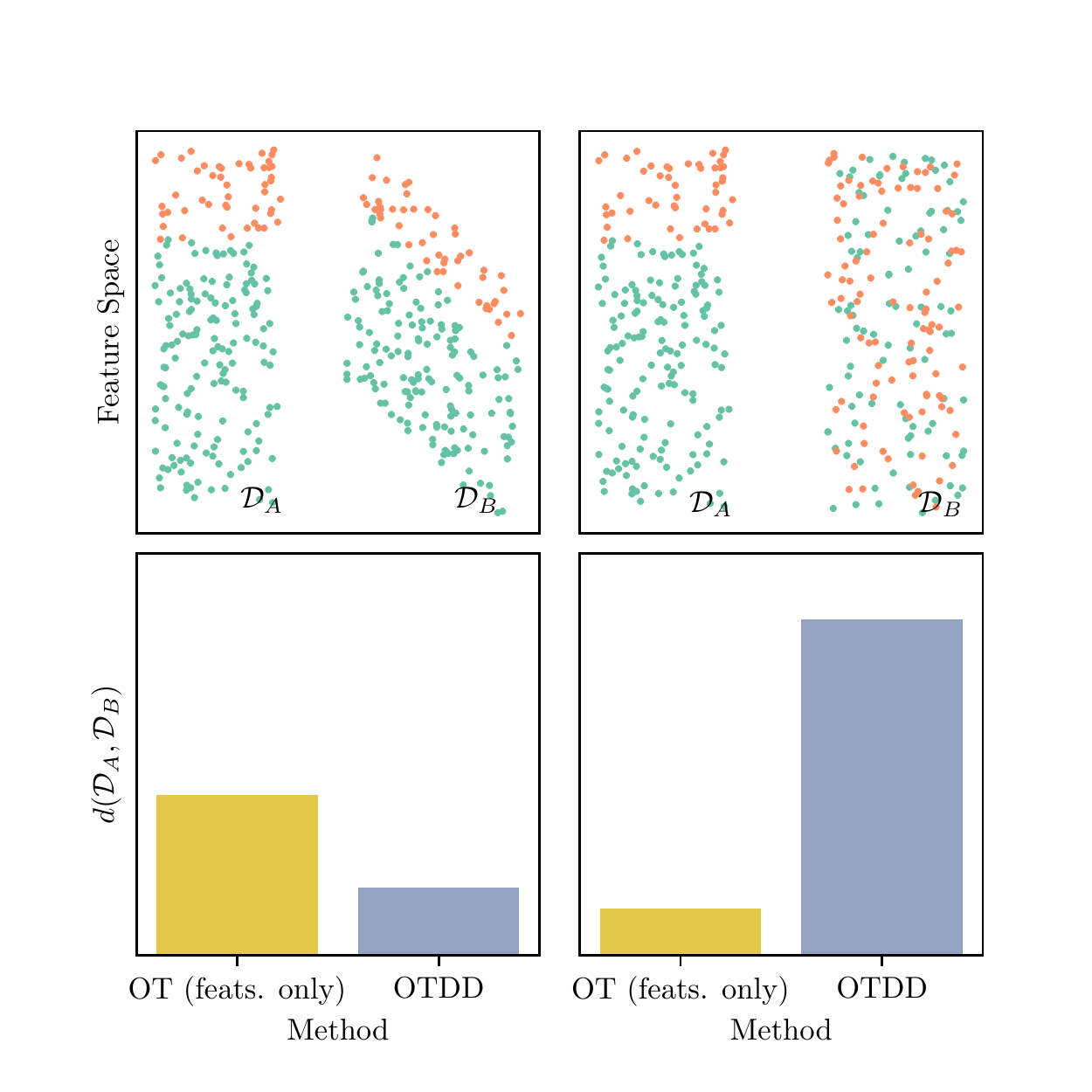}
    \caption{\textbf{The importance of labels}: the second pair of datasets are much closer than the first under the usual (label-agnostic) OT distance, while the opposite is true for our (label-aware) distance.}
    \label{fig:intuition}
\end{figure}

%% file: approach.tex
\section{Optimal Transport between Datasets}\label{sec:approach}

\begin{figure*}
    \centering
    \includegraphics[width=\linewidth, trim={0cm 0.5cm 0cm 0.4cm}, clip]{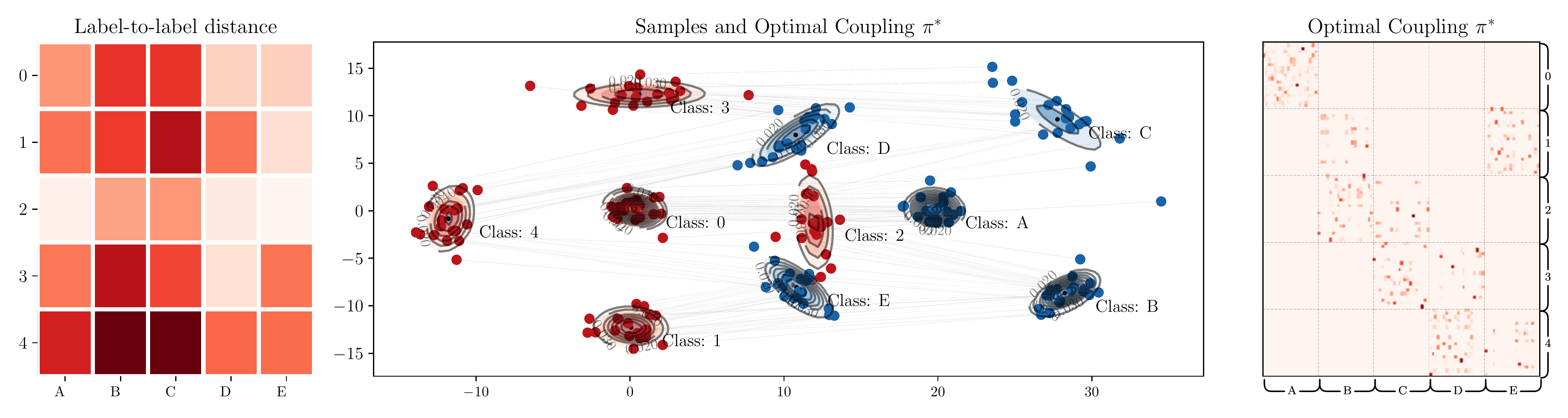}
    \caption{Our approach represents labels as distributions over features and computes Wasserstein distances between them (left). Combined with the usual metric between features, this yields a transportation cost between datasets. The optimal transport problem then characterizes the distance between them as the minimal possible cost of coupling them (optimal coupling $\pi^*$ shown on the right).}
    \label{fig:2d_gaussian_summary}
\end{figure*}

The definition of \textit{dataset} is notoriously inconsistent across the machine learning literature, sometimes referring only to features or both features and labels.
%might refer to any subset of: features, labels, losses and hypothesis classes. 
In the context of supervised learning, where the ultimate goal is to estimate predictors $f:\cX \rightarrow \cY$ (or conditional distributions $P(y\mid x)$), we define a dataset $\cD$ as a set of feature-label pairs $(x,y) \in \cX \times \cY$ over a certain feature space $\cX$ and label set $\cY$. For simplicity, we will use $z \triangleq (x,y)$ to denote these pairs, and $\cZ \triangleq \cX \times \cY$ for their underlying space. 

Henceforth, we focus on the case of classification, so $\cY$ shall be a finite set. We consider two datasets $\cD_A$ and $\cD_B$, and assume, for simplicity, that their feature spaces have the same dimensionality, but will discuss how to relax this assumption later on. On the other hand, we make no assumptions on the label sets $\cY_A$ and $\cY_B$ whatsoever. In particular, the classes these encode could be partially overlapping or related (\eg \textsc{imagenet} and \textsc{cifar-10}) or completely disjoint (\eg \textsc{cifar-10} and \mnist). Although not a formal assumption of our approach, it will be useful to think of the samples in these two datasets as being drawn from joint distributions $P_A(x,y)$ and $P_B(x,y)$. 

Given $\mathcal{D}_A = \{(x^{(i)}_A, y_A^{(i)})\}_{i=1}^n \sim P_A(x,y)$ and $\mathcal{D}_B = \{(x_B^{(j)}, y_B^{(j)})\}_{j=1}^m \sim P_B(x,y)$, our goal is to define a distance $d(\cD_A, \cD_B)$ that depends exclusively on the information contained in these datasets. The probabilistic interpretation of these collections suggests a simple-yet-proven approach: comparing these datasets by means of a statistical divergence on their joint distributions. Among many such notions, optimal transport stands out because of various characteristics described in Section~\ref{sec:background}: its direct use of the geometry of the underlying space, its characterization of distance as correspondence (which will prove to have various useful applications in this context) and the vast theory, spanning three centuries, which it is built upon.

Note, however, that direct application of OT to this setting is challenging. Indeed, problem \eqref{eq:ot_problem} requires us to define a metric on the ground space, \ie on $\cZ=\cX \times \cY$. A straightforward way to do so would be via the individual metrics in $\cX$ and $\cY$. Indeed, if $d_{\cX}, d_{\cY}$ are metrics on $\cX$ and $\cY$ respectively, then $d_{\cZ}:\cX\times \cY \rightarrow \R^+$, given as:
$$ d_{\cZ}(z,z') = \bigl ( d_{\cX}(x,x')^p + d_{\cY}(y,y')^p \bigr)^{1/p},$$
for $p\geq 1$ is a metric on $\cZ$.

In most applications, $d_{\cX}$ is readily available, \eg as the euclidean distance in the feature space. On the other hand, $d_{\cY}$ will rarely be so, particularly between labels from unrelated label sets (\eg between \texttt{cars} in one image domain and  and \texttt{dogs} in the other). If we had some prior knowledge of the label spaces, we could use it to define a notion of distance between pairs of labels. However, in the challenging\;---but common---\;case where no such knowledge is available, the only information we have about the labels is their occurrence in relation to the feature vectors $x$. Thus, we can take advantage of the fact that we have a meaningful metric in $\cX$ and use it to compare labels. Arguably, the simplest such approach is as follows. Let us define $\mathcal{N}_{\mathcal{D}}(y) := \{ x \in \cX \mid (x,y) \in \mathcal{D}\}$, \ie $\mathcal{N}_{\mathcal{D}}(y)$ is the set of feature vectors with label $y$ in dataset $\mathcal{D}$, and let $n_y$ be its cardinality. With this, a distance between two labels $y$ and $y'$ can be defined as the distance between the centroids of their associated feature vector collections:
\begin{equation}
	d(y,y') =  d_{\cX}\biggl(\frac{1}{n_y}\sum_{x \in \mathcal{N}_{\mathcal{D}}(y)} x, \medspace \frac{1}{n_{y'}}\sum_{x \in \mathcal{N}_{\mathcal{D}}(y')} x \biggr).
\end{equation}
Although appealing for its simplicity, representing the collections $\mathcal{N}_{\mathcal{D}}(y)$ only through their mean is too simplistic for real datasets. Ideally, we would like to represent labels through the \textit{actual distribution} over the feature space that they define, namely, by means of the map $y \mapsto \alpha_y(X) \triangleq P(X \mid Y=y)$, of which $\mathcal{N}_{\mathcal{D}}(y)$ can be understood as a finite sample. If we use this representation, defining a distance between labels boils down to choosing a statistical divergence between their associated distributions. Once more, there are many possible choices for this distance, but ---yet again--- we argue that an OT is an ideal choice, since the notion of divergence we seek should: (i) provide a valid metric, (ii) be computable from finite samples, which is crucial since the distributions $\alpha_y$ are not available in analytic form, and (iii) be able to deal with sparsely-supported distributions, all of which OT satisfies.

The approach described so far \textit{grounds} the comparison of the $\alpha_y$ distributions to the feature space $\cX$, so we can simply use $d_{\cX}^p$ as the optimal transport cost, leading to a p-Wasserstein distance between labels: $\text{W}_p^p(\alpha_y, \alpha_{y'})$, and in turn, to the following distance between feature-label pairs:
\begin{equation}\label{eq:ottask_metric}
	d_{\cZ}\bigl((x,y), (x',y') \bigr) \triangleq \bigl( d_{\cX}(x,x')^p  + \text{W}_p^p(\alpha_y, \alpha_{y'}) \bigr)^{\frac{1}{p}}.
\end{equation}

This gives us a point-wise notion of distance in $\cZ$, but we ultimately seek a distance \textit{between distributions} over this space, \ie between joint distributions $P(x,y)$. Optimal transport allows us to lift the \emph{ground} (\ie point-wise) metric defined above into a distance between measures:

\begin{equation}\label{eq:ottask_general_distance}
	d_{\OT}(\cD_A,\cD_B) = \min_{\pi \in \Pi(\alpha, \beta)}\int_{\cZ \times \cZ} d_{\cZ}(z,z') \pi(z,z').
\end{equation}

The following result, an immediate consequence of the discussion above, states that Eq.~\eqref{eq:ottask_general_distance} is a proper distance -- the Optimal Transport Dataset Distance (\textsc{otdd}).

\begin{proposition}\label{prop:distance_is_distance}
		$d_{\OT}(\cD_A, \cD_B)$ defines a valid metric on  $\cP(\cX \times \cP(\cX))$ the space of measures over feature and label-distribution pairs.
\end{proposition}

It remains to describe how the distributions $\alpha_y$ are to be represented. A flexible non-parametric approach would be to treat the samples in $\mathcal{N}_{\mathcal{D}}(y)$ as support points of a uniform empirical measure, \ie $\alpha_y = \sum_{\x^{(i)} \in \mathcal{N}_{\cD}(y)} \frac{1}{n_y}\delta_{\x^{(i)}}$, as described in Section~\ref{sec:background}. The main downside of this approach is that each evaluation of \eqref{eq:ottask_metric} involves solving an optimization problem, which could be prohibitive. Indeed, in Section~\ref{sec:complexity_general} we show that for datasets of size $n$, this approach has worst-case $O(n^5 \log n)$ complexity.

Instead, we propose an alternative representation of the $\alpha_y$ as Gaussian distributions, which leads to a simple yet tractable realization of the general dataset distance \eqref{eq:ottask_general_distance}. Formally, let us denote by $\hat{\mu}_y \in \R^d$ and $\hat{\Sigma}_y \in \R_+^{d\times d}$ the sample mean and covariance matrix associated with label $y$ (through its feature neighborhood $\mathcal{N}_{\mathcal{D}}(y))$, that is:
\begin{equation*}
	\hat{\mu}_y \triangleq \frac{1}{n_y}\sum_{x \in \mathcal{N}_{\mathcal{D}}(y)} x; \medspace
	\hat{\Sigma}_y  \triangleq  \frac{1}{n_y}\sum_{x \in \mathcal{N}_{\mathcal{D}}(y)}(x - \hat{\mu}_y)^\top(x-\hat{\mu}_y).
\end{equation*}
With this, we model each label-feature distribution $\alpha_y$ as a Gaussian Distribution $\mathcal{N}(\hat{\mu}_y, \hat{\Sigma}_y)$ whose parameters are the sample mean and covariance of $\mathcal{N}_{\mathcal{D}}(y)$. 

The main motivation behind this choice is that the 2-Wasserstein distance between Gaussian distributions $\cN(\mu_{\alpha}, \Sigma_{\alpha})$ and $\cN(\mu_{\beta}, \Sigma_{\beta})$ has as an analytic form:
\begin{equation}\label{eq:gaussian_wasserstein}
	\text{W}_2^2(\alpha, \beta) = \| \mu_{\alpha}\!-\!\mu_{\beta} \|_2^2 + \tr(\Sigma_\alpha \!+\!\Sigma_{\beta}\! -2(\Sigma_{\alpha}^{\frac{1}{2}}\Sigma_{\beta}\Sigma_{\alpha}^{\frac{1}{2}})^{\frac{1}{2}})
\end{equation}
where $\Sigma^{\frac{1}{2}}$ denotes the matrix square root. Furthermore, whenever $\Sigma_{\alpha}$ and $\Sigma_{\beta}$ commute, this further simplifies to
\begin{equation}\label{eq:gaussian_wasserstein_commute}
		\text{W}_2^2(\alpha, \beta) = \| \mu_{\alpha} - \mu_{\beta} \|_2^2 +  \| \Sigma^\frac{1}{2}_\alpha - \Sigma_{\beta}^\frac{1}{2} \|_{F}^2.
\end{equation}
When using Eq.~\eqref{eq:gaussian_wasserstein} in the point-wise distance \eqref{eq:ottask_metric}, we denote the resulting distance \eqref{eq:ottask_general_distance} by $d_{\OT\mhyphen\mathcal{N}}$. 

Representing label-defined distributions as Gaussians might seem like a heuristic choice driven only by algebraic convenience. However, the following result, a consequence of a bound by \citet{gelbrich1990formula}, shows that this approximation lower-bounds the distance that would be obtained had it been computed using the label distances on the \textit{true} distributions (regardless of their form):
\begin{proposition}\label{prop:extended_gelbrich}
    For any two datasets $\cD_A, \cD_B$, we have:
    \begin{equation}
        d_{\OT\mhyphen\cN}(\cD_A, \cD_B) \leq d_{\OT}(\cD_A, \cD_B)
    \end{equation}
    Furthermore, if the label distributions $\alpha_y$ are all Gaussian or elliptical, these quantities are equal, \ie $d_{\OT\mhyphen\cN}$ is exact. 
\end{proposition}

An illustration of the OTDD in a synthetic dataset summarizing its main characteristics is shown in Figure~\ref{fig:2d_gaussian_summary}.

%% file: computation.tex
\section{Computational Considerations}\label{sec:computation}
\vspace{-0.05cm}
Since our goal in this work is to use the proposed dataset distance as a tool for tasks like transfer learning in realistic (\ie large) machine learning datasets, scalability is crucial. Indeed, most compelling use cases of \textit{any} notion of distance between datasets will involve computing it repeatedly on very large samples. 

\vspace{-0.05cm}
While estimation of Wasserstein ---and more generally, optimal transport--- distances is known to be computationally expensive in general, in Section~\ref{sec:background} we briefly discussed how entropy regularization can be used to trade-off accuracy for runtime. Recall that both the general and Gaussian versions of the dataset distance proposed in Section~\ref{sec:approach} involve solving optimal transport problems (though the latter, owing the closed form solution of subproblem~\eqref{eq:gaussian_wasserstein}, only requires optimization for the global problem). Therefore, both of these distances benefit from approximate OT solvers. 

\vspace{-0.05cm}
But further speed-ups are possible. For $d_{\OT\mhyphen\cN}$, a simple and fast implementation can be obtained if (i) the metric in $\cX$ coincides with the ground metric in the transport problem on $\cY$, and (ii) all covariance matrices commute. While (ii) will rarely occur in practice, one could use a diagonal approximation to the covariance, or with milder assumptions, simultaneous matrix diagonalization \citep{delathauwer2003simultaneous}. In either case, using the simplification in \eqref{eq:gaussian_wasserstein_commute}, the pointwise distance $d(z,z')$ can be computed by creating augmented representations of each dataset, whereby each pair $(x,y)$ is represented as a stacked vector $\tilde{x}:=[x;\mu_y;\text{vec}(\Sigma_y^{1/2})]$ for the corresponding label mean and covariance. Then,
$ \| \tilde{x} - \tilde{x}'\|_2^2 = d_{\cZ}(x,y; x',y')^2 $
for $d_{\cZ}$ as defined in Eq.~\eqref{eq:ottask_metric}. Therefore, in this case the OTDD can be immediately computed using an off-the-shelf OT solver on these augmented datasets. While this approach is appealing computationally, here instead we focus on a exact version that does not require diagonal or commuting covariance approximations, and leave empirical evaluation of this approximate approach for future work.

\begin{figure*}[th]
    \centering
    \includegraphics[width=0.32\linewidth, trim={0.2cm 0.2cm 0 0}, clip]{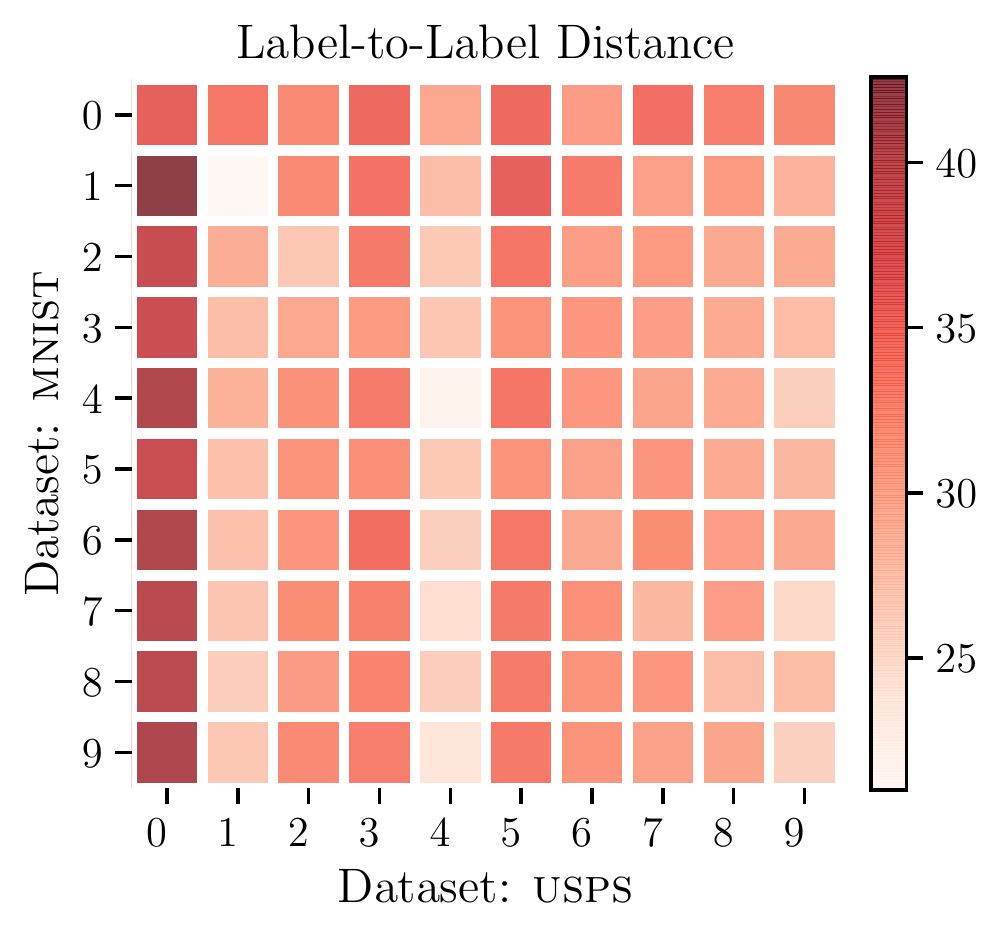}%
    \includegraphics[width=0.29\linewidth]{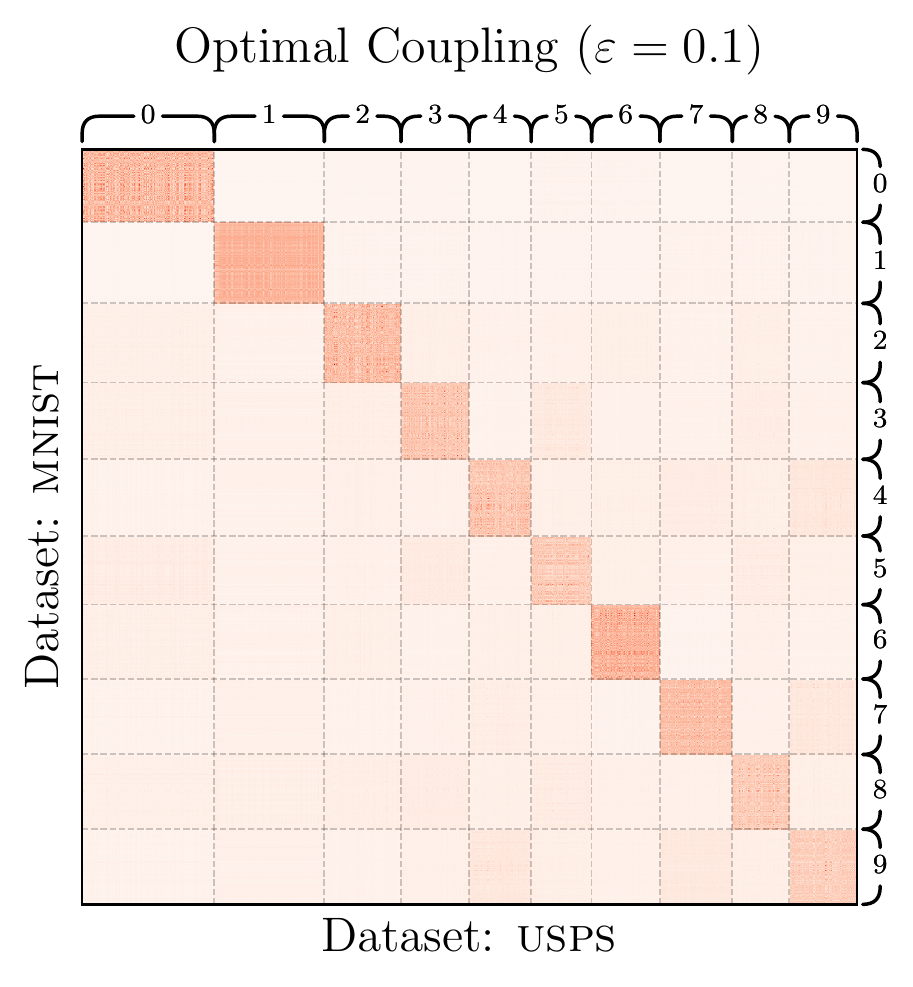}%
    \includegraphics[width=0.29\linewidth]{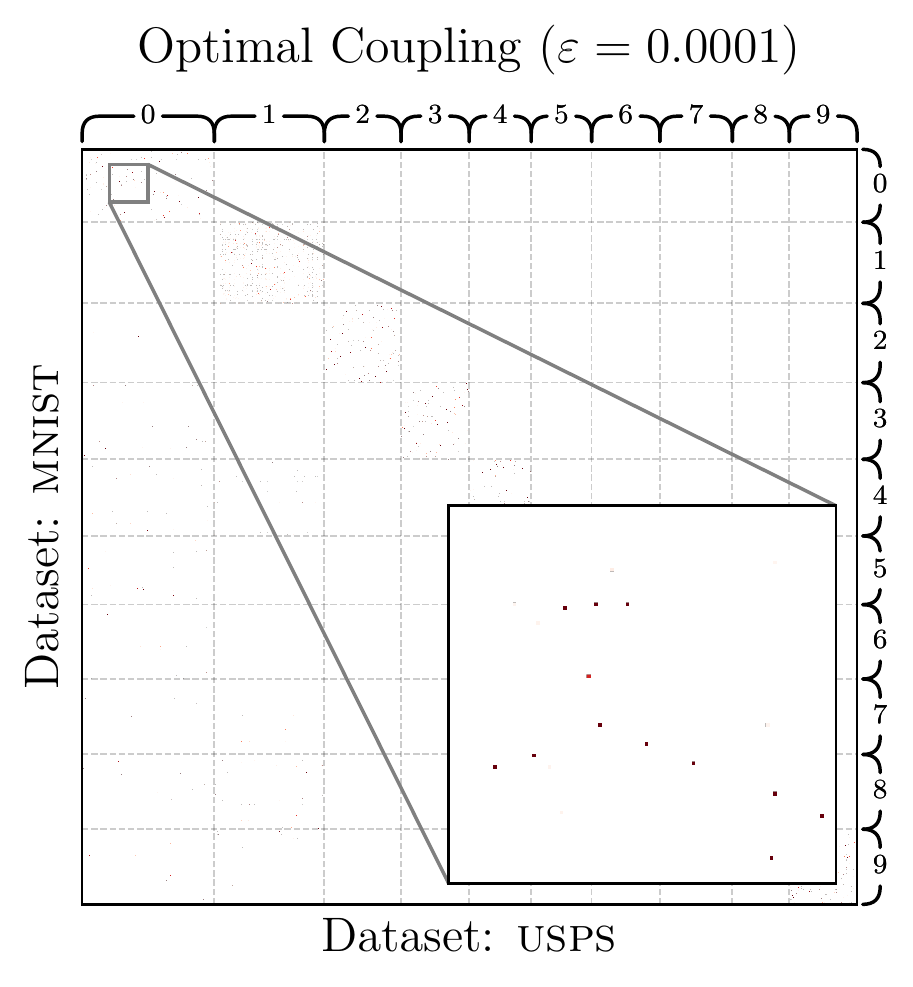}%
    \caption{\textbf{Dataset Distance between \mnist and \usps}. \textbf{Left}: The label Wasserstein distances ---computed without knowledge of the relation between labels across domains--- recover expected relations between classes in the two domains. \textbf{Center/Right}: The optimal coupling $\pi^*$ for different regularization levels exhibits a block-diagonal structure, indicating class-coherent matches across domains. 
    }
    \label{fig:mnist_usps}
\end{figure*}

\vspace{-0.05cm}
The steps we propose next are motivated by the observation that, unlike traditional OT distances for which the cost of computing pair-wise distance is negligible compared to the complexity of the optimization routine, in our case the latter dominates, since it involves computing multiple OT distances itself. In order to speed up computation, we first precompute and store in memory all label-to-label pairwise distances $d(\alpha_y, \alpha_{y'})$, and retrieve them on-demand during the optimization of the global OT problem. 

\vspace{-0.05cm}
For $d_{\OT\mhyphen\cN}$, computing the label-to-label distances  $d(\mathcal{N}(\hat{\mu}_y, \hat{\Sigma}_y),  \mathcal{N}(\hat{\mu}_{y'}, \hat{\Sigma}_{y'}))$ is dominated by the cost of computing matrix square roots, which if done exactly involves a full eigendecomposition. Instead, it can be computed approximately using the Newton-Schulz iterative method \citep{higham2008functions, muzellec2018generalizing}. Besides runtime, loading all examples of a given class to memory (to compute means and covariances) might be infeasible for large datasets (especially if running on GPU), so we instead use a two-pass stable online batch algorithm to compute these statistics \citep{chan1983algorithms}. 

\vspace{-0.05cm}
The following result summarizes the time complexity of our two distances and sheds light on the trade-off between precision and efficiency they provide.

\vspace{-0.1cm}
\begin{theorem}\label{thm:complexity}
    For datasets of size $n$ and $m$, with $p$ and $q$ classes, dimension $d$, and maximum class size $\mathfrak{n}$, both $d_{\OT}$ and $d_{\OT\mhyphen\cN}$ incur in a cost of  
    $O(nm\log(\max\{n,m\}) \tau^{-3})$ for solving the global OT problem $\tau$-approximately, while the worst-case complexity for computing the label-to-label pairwise distances \eqref{eq:ottask_metric} is $O\bigl(nm (d + \mathfrak{n}^3\log \mathfrak{n} + d\mathfrak{n}^2)\bigr)$ for $d_{\OT}$ and $O\bigl(nmd+pqd^3+d^2\mathfrak{n}(p+q)\bigr)$ for $d_{\OT\mhyphen\cN}$.
\end{theorem}
\vspace{-0.1cm}

 In most practical applications, the cost of computing pairwise distances will dominate, making $d_{\OT\mhyphen\cN}$ superior. For example, if $n=m$ and the largest class size is $O(n)$, this step becomes $O(n^5\log n)$ \;---prohibitive for all but toy datasets---\; for $d_{\OT}$ but only $O(n^2d+d^3)$ for $d_{\OT\mhyphen\cN}$. 

%% file: experiments.tex
\section{Experiments}\label{sec:experiments}

\subsection{Dataset Selection for Transfer Learning}

A driving motivation for proposing a dataset distance was to provide a learning-free criterion on which to select a source dataset for transfer learning. In this section, we put this hypothesis to test on a simple domain adaptation setting on \mnist \citep{lecun2010mnist} and three of its extensions: \textsc{fashion-mnist} \citep{xiao2017fashion-mnist}, \textsc{kmnist} \citep{clanuwat2018deep} and the \texttt{letters} split of \textsc{emnist} \citep{cohen2017emnist}, in addition to \usps. All datasets consist of 10 classes, except \textsc{emnist}, for which the selected split has 26 classes. Throughout this section, we use a simple LeNet-5 neural network (two convolutional layers, three fully conntected ones) with ReLU activations. When carrying out adaptation, we freeze the convolutional layers and fine-tune only the top three layers. 

\begin{figure}
    \centering
    \includegraphics[width=0.75\linewidth, trim={.5cm 3.5cm 0.5 1.5cm}, clip]{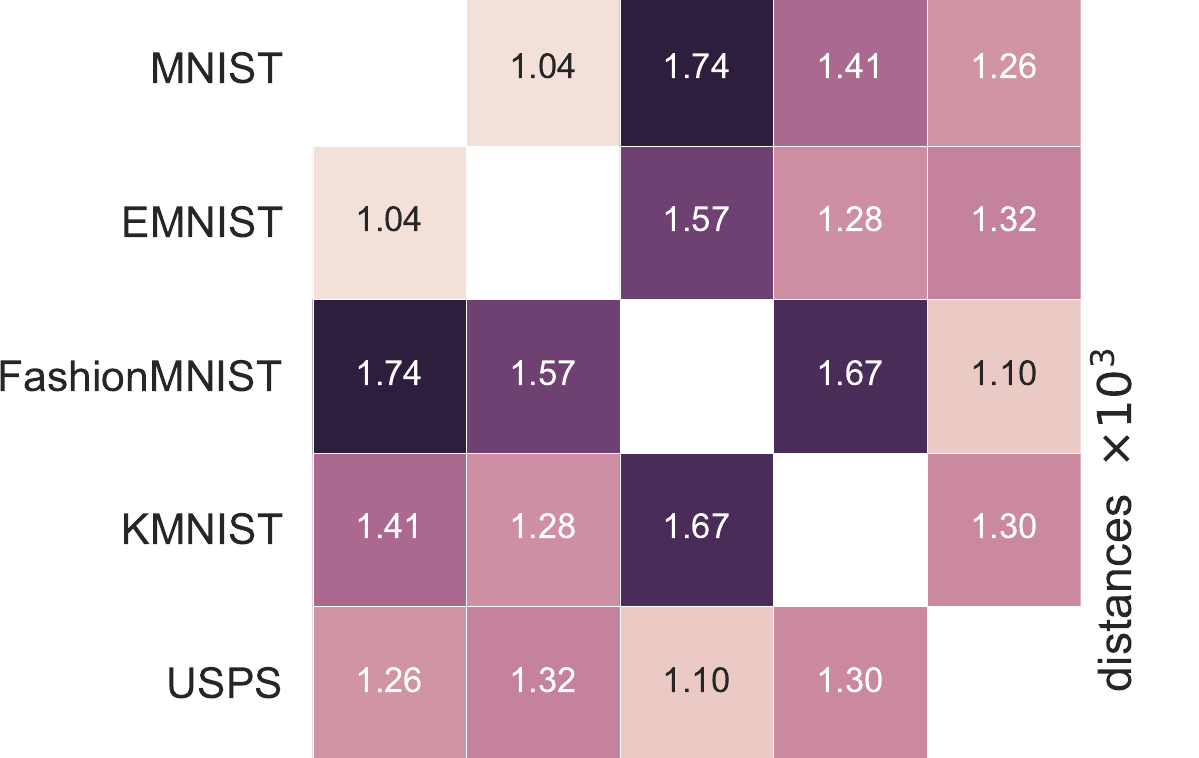}
    \caption{Pairwise OT Distances for *\textsc{nist}+\textsc{usps} datasets.}\label{fig:nist_dist_heatmap}
    \vspace{-0.2cm}
\end{figure}

\vspace{-0.05cm}
We first compute all pairwise OTDD distances (Fig~\ref{fig:nist_dist_heatmap}). For the example of $d_{\OT\mhyphen\cN}(\mnist,\usps)$, Figure~\ref{fig:mnist_usps} illustrates two key components of the computation of the distance: the label-to-label distances (left) and the optimal coupling $\pi^*$ obtained for two choices of entropy regularization parameter $\varepsilon$ (center, right). The diagonal elements of the first plot (\ie distances between corresponding digit classes) are overall relatively smaller than off-diagonal elements. Interestingly, the \texttt{0} class of \usps appears remarkably far from \textit{all} \mnist digits under this metric. On the other hand, most correspondences lie along the (block) diagonal of $\pi^*$, which shows the dataset distance is able to infer class-coherent correspondences across them. 

\vspace{-0.05cm}
We test the robustness of the distance by computing it repeatedly for varying sample sizes. The results (Fig.~\ref{fig:mnist_usps_robustness}, Appendix~\ref{sec:robustness}) show that the distance converges towards a fixed value as sample sizes grow, but interestingly, small sample sizes for \usps lead to wider variability, suggesting that this dataset itself is more heterogeneous than \mnist. 

Despite both consisting of digits, \mnist and \usps are not the closest among these datasets according to the OTDD, as Figure~\ref{fig:nist_dist_heatmap} shows. The closest pair is instead (\mnist, \textsc{emnist}), while \textsc{fashion-mnist} appears comparatively far from all others, particularly \mnist. 

\begin{figure}
    \centering
    \includegraphics[width=\linewidth, trim={0.1cm 0.1cm 0.1cm 0.1cm},clip]{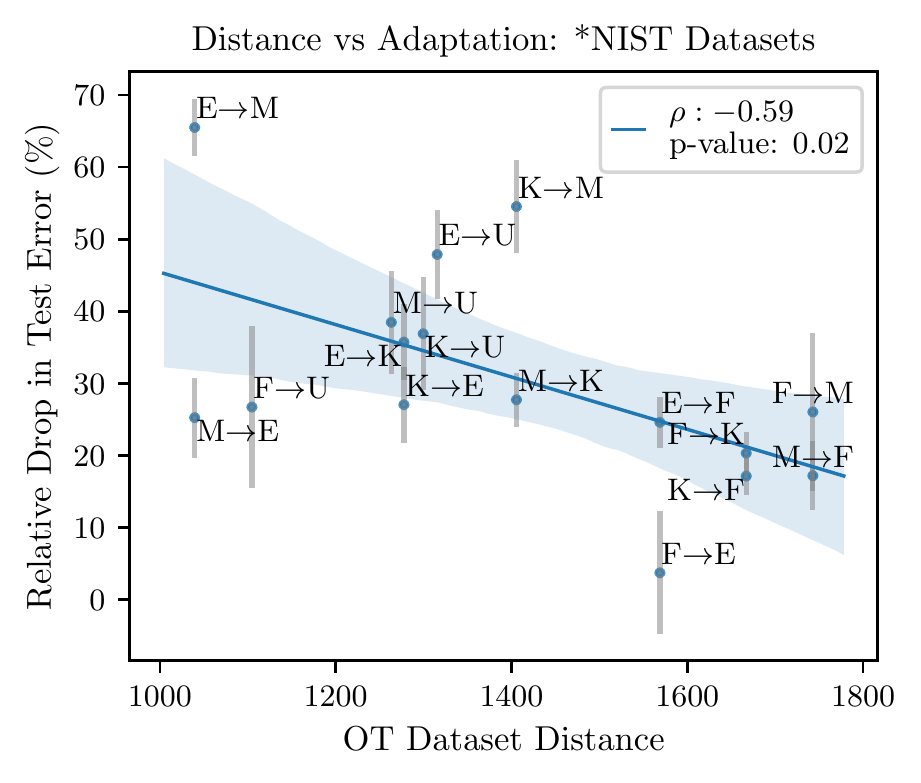}
    \caption{Dataset distance vs.~adaptation for *NIST datasets (M: \textsc{mnist}, E: \textsc{emnist}, K: \textsc{kmnist}, F: \textsc{fashion-mnist}, U: \textsc{usps}). The error bars correspond to $\pm$1 s.d. over 10 repetitions.
    }\label{fig:nist_dist_adap}
\end{figure}

Next, we compare these distances against the \textit{transferability} between datasets, \ie the gain in performance from using a model pertrained on the source domain and fine-tuning it on the target domain. To make these numbers comparable across adaptation pairs which involve datasets of very different hardness, we define the transferability $\cT$ of a source domain $\cD_S$ to a target domain $\cD_T$ as the relative decrease in classification error when doing adaptation compared to training only on the target domain, \ie 

\begin{equation*}\label{eq:relative_error}
    \cT(\cD_S\!\to\!\cD_T) = 100\times\frac{\mathrm{error}(\cD_S\!\to\!\cD_T) - \mathrm{error}(\cD_T)}{\mathrm{error}(\cD_T)}.
\end{equation*}

We run the adaptation task 10 times with different random seeds for each pair of datasets, and compare $\cT$ against their distance.  The strong significant correlation between these (Fig.~\ref{fig:nist_dist_adap}) shows that the OTDD is highly predictive of transferability across these datasets. In particular, \textsc{emnist} led to the best adaptation to \mnist, justifying the\;---initially counter-intuitive---\;value of the OTDD.

\begin{figure}
    \centering
    \includegraphics[width=\linewidth, trim={0.1cm 0.1cm 0.1cm 0.1cm}, clip]{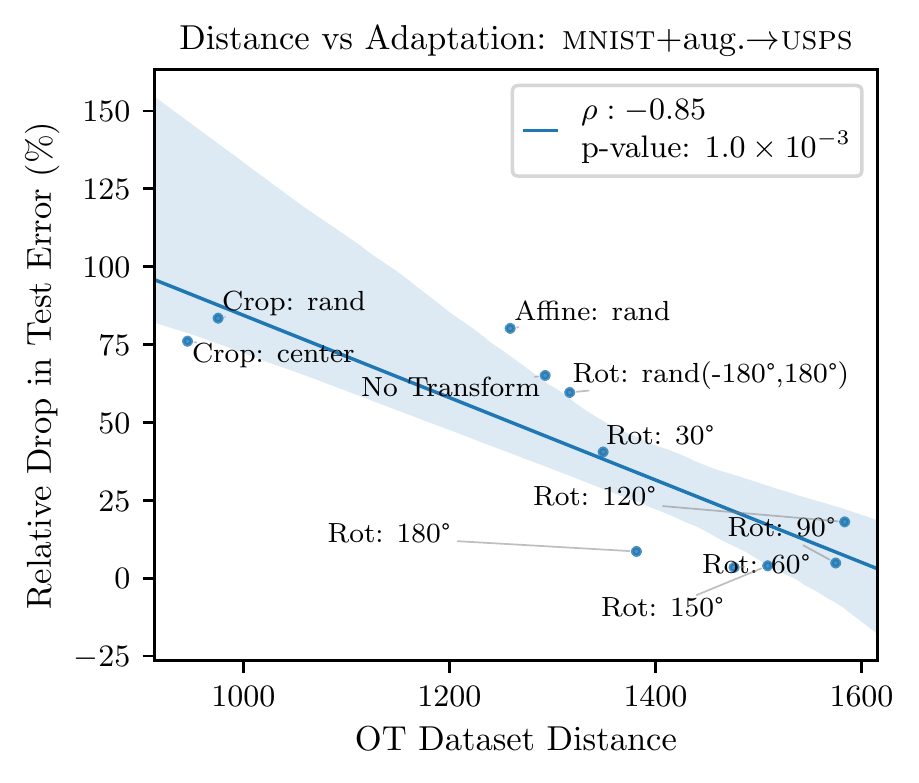}
    \caption{Dataset distance vs.~adaptation between \mnist with various transformations applied to it and \usps. While cropping the \mnist digits leads to better adaptation, rotating them degrades it, both in agreement with the dataset distance.
    }\label{fig:mnist_usps_augmentation}
\end{figure}

\subsection{Distance-Driven Data Augmentation}

Data augmentation\;---\ie applying carefully chosen transformations on a dataset to enhance its quality and diversity---\;is another key aspect of transfer learning that has substantial empirical effect on the quality of the transferred model yet lacks principled guidelines. Here, we investigate if the OTDD could be used to compare and select among possible augmentations. 

\vspace{-0.05cm}
For a fixed source-target dataset pair, we generate replicas of the source data with various transformations applied to it, compute their distance to the target dataset, and compare against the transferability as before. We present results for a small-scale (\mnist$\to$\usps) and a larger-scale (Tiny-ImageNet$\to$\textsc{cifar}-10) setting. The transformations we use on \mnist consist of rotations by a fixed degree $[30^{\circ},\dots,180^{\circ}]$, random rotations $(-180^{\circ},180^{\circ})$, random affine transformations, center-crops and random crops. For Tiny-ImageNet we randomly vary brightness, contrast, hue and saturation. The models use are respectively the LeNet-5 and a ResNet-50 (training details provided in Appendix~\ref{sec:training_details}). 

\begin{figure}
    \centering
    \includegraphics[width=\linewidth, trim={0 0 0 0}, clip]{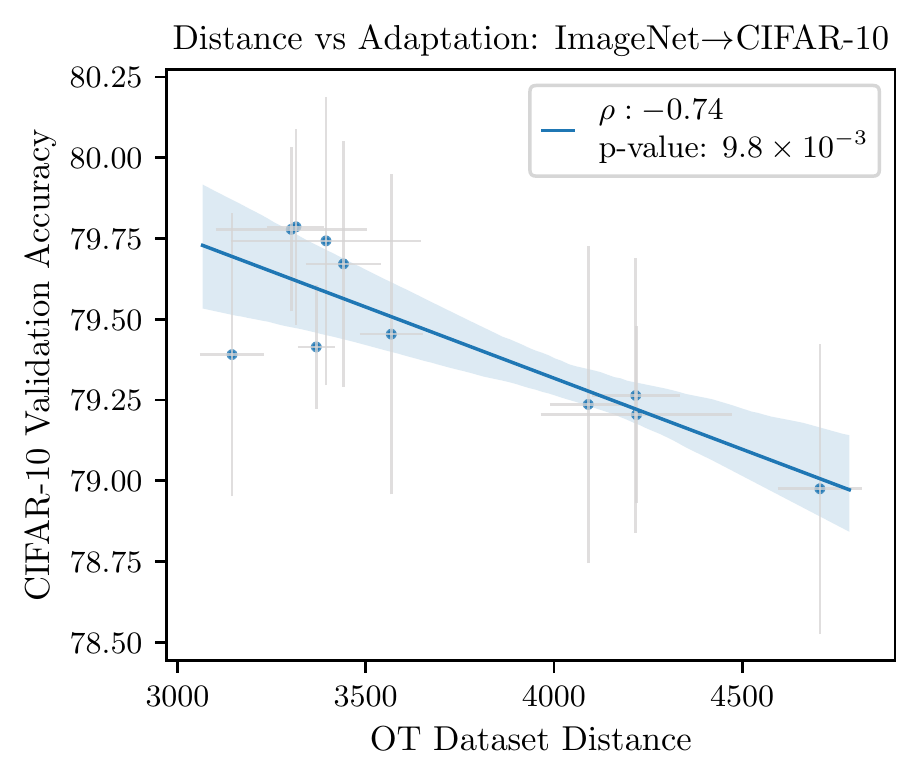}
    \caption{Dataset distance vs.~adaptation between Tiny-ImageNet with various transformations (source) and \textsc{cifar-10} (target).
    }\label{fig:imagenet_cifar_augmentation}
\end{figure}

\vspace{-0.05cm}
The results in both of these settings (Figures~\ref{fig:mnist_usps_augmentation} and \ref{fig:imagenet_cifar_augmentation}) show, again, a strong significant correlation between these two. A reader familiar with the \mnist and \usps datasets will not be surprised by the fact that cropping images from the former leads to substantially better performance on the latter, while most rotations degrade transferability.

\vspace{-0.05cm}
\subsection{Transfer Learning for Text Classification}\vspace{-0.05cm}
Natural Language Processing (NLP) is of the areas where large-scale transfer learning has had the most profound impact over the past few years, in part driven by the availability of off-the-shelf large language-models pretrained on massive amounts of the data \citep{peters2018deep, devlin2019bert, radford2019better}. 

While natural language inherently lacks the fixed-size continuous vector representation required by our framework to compute pointwise distances, we can take advantage of precisely these pretrained models to embed sentences in vector space, furnishing them with a rich geometry. In our experiments, we first embed every sentence of every dataset using the (base) \textsc{bert} model \citep{devlin2019bert},\footnote{Using the \texttt{sentence\_transfomers} library.} and then compute OTDD on these embedded datasets.

Here, we focus on the problem of sentence classification, and consider the following datasets\footnote{Available via the \texttt{torchtext} library.} by \citet{zhang2015character-level}: \textsc{ag news} ($\mathrm{ag}$), \textsc{dbpedia} ($\mathrm{db}$), \textsc{yelp reviews} with 5-way classification ($\mathrm{yl}_5$) and binary polarity ($\mathrm{yl}_+$) label encodings, \textsc{amazon reviews} with 5-way classification ($\mathrm{am}_5$) and binary polarity ($\mathrm{am}_+$) label encodings, and \textsc{yahoo answers} ($\mathrm{yh}$). We provide details for all these datasets in the Appendix. 

\begin{figure}
    \centering
    \includegraphics[width=\linewidth,trim={0.1cm 0.1cm 0.1cm 0.1cm}, clip]{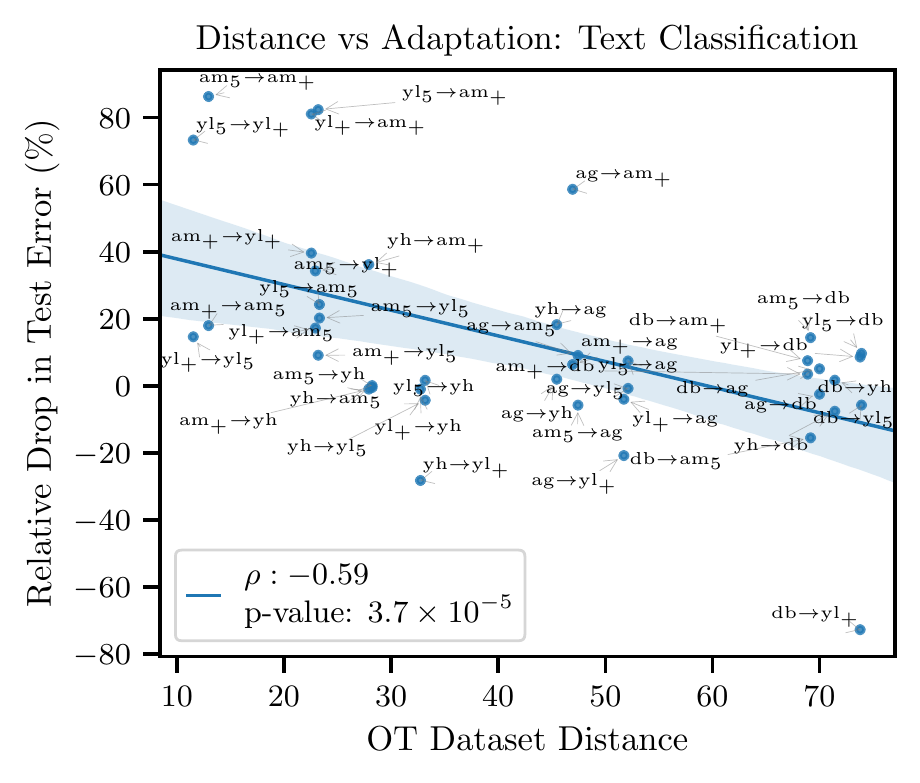}    
    \caption{Distance vs.~adaptation for text classification datasets (see main text for key), with sentence embedding via \textsc{bert}.
    }\label{fig:text_classification}    
\end{figure}

As before, we simulate a challenging adaptation setting by keeping only 100 examples per target class. For every pair of datasets, we first fine-tune the \textsc{bert} model using the entirety of the source domain data, after which we fine-tune and evaluate on the target domain. Figure~\ref{fig:text_classification} shows that the OT dataset distance is highly correlated with transferability in this setting too. Interestingly, adaptation often leads to drastic degradation of performance in this case, which suggests that off-the-shelf \textsc{bert} is on its own powerful and flexible enough to initialize many of these tasks, and therefore choosing the wrong domain for initial training might destroy some of that information. 

%% file: conclusion.tex
\section{Discussion}\label{sec:discussion}

We have shown that the notion of distance between datasets proposed in this work is scalable and flexible enough to be used in realistic transfer learning scenarios, all the while offering appealing theoretical properties, interpretable comparisons and requiring minimal assumptions on the underlying datasets. 

There are many natural extensions of this framework. Here we assumed that the datasets where defined on feature spaces of the same dimension, but one could instead leverage a relational notion such as the Gromov-Wasserstein distance \citep{memoli2011gromov} to compute the distance between datasets whose features and not directly comparable. On the other hand, our efficient implementation relies on modeling groups of points with the same label as Gaussian distributions. This could naturally be extended to more general distributions for which the Wasserstein distance either has an analytic solution or at least can be computed efficiently, such as elliptic distributions \citep{muzellec2018generalizing}, Gaussian mixture models \citep{delon2019wasserstein}, certain Gaussian Processes \citep{mallasto2017learning}, or tree metrics \citep{le2019tree-sliced}. 

In this work, we purposely excluded two key aspects of any learning task from our notion of distance: the loss function and the predictor function class. While we posit that it is crucial to have a notion of distance that is independent of these choices, it is nevertheless appealing to ask whether our distance could be extended to take those into account, ideally involving minimal training. Exploring different avenues to inject such information into this framework will be the focus of our future work.

%% file: appendix.tex
\appendix

\section{Proof of Proposition~\ref{prop:distance_is_distance}}

Whenever the cost function used in the of optimal transport problem is a metric in a given space $\cX$, the optimal transport problem is a distance (the Wasserstein distance) on $\mathcal{P}(\cX)$ \citep[Chapter~6]{villani2008optimal}. Therefore, it suffices to show that the cost function $d_{\cZ}$ defined in Eq.~\eqref{eq:ottask_metric} is indeed a distance. Clearly, it is symmetric because both $d_{\cX}$ and $\W_p$ are. In addition, since both of these are distances:
\begin{align*}
    d_{\cZ}(z,z') = 0 &\Leftrightarrow d_{\cX}(x,x')=0 \land \W_p(\alpha_y, \alpha_y')=0 \\
    &\Leftrightarrow x=x', \medspace \alpha_y = \alpha_y' \\
    &\Leftrightarrow z=z'
\end{align*}
Finally, we have that
\begin{align*}
    d_{\cZ}(z_1, z_3) &= \bigl(d_{\cX}(x_1,x_3)^p +  \W_p(\alpha_{y_1}, \alpha_{y_3})^p \bigr)^{\frac{1}{p}} \\
    & \leq \bigl(d_{\cX}(x_1,x_2)^p+ d_{\cX}(x_2,x_3)^p + \\
    &\medspace  \W_p(\alpha_{y_1}, \alpha_{y_2})^p +  \W_p(\alpha_{y_2}, \alpha_{y_3})^p \bigr)^{\frac{1}{p}} \\
    &=  \bigl( d_{\cZ}(z_1, z_2)^p + d_{\cZ}(z_2, z_3)^p \bigr)^{\frac{1}{p}} \\
    &= d_{\cZ}(z_1, z_2) + d_{\cZ}(z_2, z_3)
\end{align*}
where the last step is an application of Minkowski's inequality. Hence, $d_{\cZ}$ satisfies the triangle inequality, and therefore it is a metric on $\cZ = \cX \times \cP(\cX)$. We therefore conclude that the value of the optimal transport \eqref{eq:ottask_general_distance} that uses this metric as a cost function is a distance itself. \qed

\section{Proof of Proposition~\ref{prop:extended_gelbrich}}

Our proof relies directly on a well-known bound for the 2-Wasserstein distance between distributions by \citep{gelbrich1990formula}:

\begin{lemma}[Gelbrich bound]\label{lemma:gelbrich}
    Suppose $\alpha,\beta \in \cP(\R^d)$ are any two measures with mean vectors $\mu_\alpha,\mu_\beta \in \R^d$ and covariance matrices $\Sigma_\alpha,\Sigma_\beta \in \mathbb{S}_{+}^d$ respectively. Then, 
    \begin{equation}
        \W_2^2\bigl(\cN(\mu_\alpha,\Sigma_\alpha), \cN(\mu_\beta, \Sigma_\beta)) \leq \W_2^2(\alpha, \beta) 
    \end{equation}
    where $\W_2^2\bigl(\cN(\mu_\alpha,\Sigma_\alpha), \cN(\mu_\beta, \Sigma_\beta))$ is as in Eq.~\eqref{eq:gaussian_wasserstein}.
\end{lemma}

In the notation of Section 3, Lemma~\ref{lemma:gelbrich} implies that for every feature-label pairs $z=(x,y)$ and $z'=(x',y')$, we have:
\begin{multline}
         d_{\cX}(x,x') + \W_2^2\bigl(\cN(\mu_y,\Sigma_y),\cN(\mu_{y'},\Sigma_{y'})) \\ \leq d_{\cX}(x,x') + \W_2^2(\alpha_y, \alpha_{y'}) 
\end{multline}

Therefore:
\begin{equation}
    \int d_{\cZ}(z,z') \dif \pi \leq \int d_{\cZ}(z,z') \dif \pi 
\end{equation}
for every coupling $\pi \in \Pi(\alpha, \beta)$. In particular, for the minimizing $\pi^*$, we obtain that 
\begin{equation}\label{eq:dist_ineq_proof}
        d_{OT}(\cD_A, \cD_B; \cN) \leq d_{OT}(\cD_A, \cD_B)
\end{equation}

Clearly, Gelbrich's bound holds with equality when $\alpha$ and $\beta$ are indeed Gaussian. More generally, equality is attained for elliptical distributions with the same density generator \citep{kuhn2019wasserstein}). This immediately implies equality of the two quantities in equation \eqref{eq:dist_ineq_proof} in that case. \qed

\begin{table*}[ht!]
  \scriptsize
  \centering
%   \ra{1.28}
      \flushleft
         \resizebox{\linewidth}{!}{%
		\begin{tabular}{r c c c c c} 
			\toprule
			 Dataset & Input Dimension & Number of Classes & Train Examples & Test Examples & Source \\
			\midrule
            \textsc{usps} & $16\times 16^{\ast}$ & $10$ & $7291$ & $2007$ & \citep{hull1994database} \\
            \textsc{mnist} & $28\times 28$ & $10$ & $60$K & $10$K & \citep{lecun2010mnist}\\
            \textsc{emnist} (letters) & $28\times 28$ & 26 & $145$K & $10$K & \citep{cohen2017emnist}\\
            \textsc{kmnist} & $28\times 28$ & $10$ & $60$K & $10$K & \citep{clanuwat2018deep}\\
            \textsc{fashion-mnist} & $28\times 28$ & 10 &$60$K & $10$K & \citep{xiao2017fashion-mnist}\\
            \midrule
            \textsc{Tiny-ImageNet} & $64\times 64^{\ddag}$ & $200$ & $100$K & $10$K & \citep{deng2009imagenet} \\
            \textsc{cifar-10} & $32\times 32$ & $10$ & $50$K & $10$K & \citep{krizhevsky2009learning} \\
            \midrule
            \textsc{ag-news} & $768^{\dag}$ & $4$ & $120$K & $7.6$K & \citep{zhang2015character-level}\\
            \textsc{DBpedia} & $768^{\dag}$ & $14$ & $560$K & $70$K & \citep{zhang2015character-level}\\
            \textsc{YelpReview} (Polarity) & $768^{\dag}$ & $2$ & $560$K & $38$K & \citep{zhang2015character-level}\\
            \textsc{YelpReview} (Full Scale)& $768^{\dag}$ & $5$ & $650$K & $50$K & \citep{zhang2015character-level}\\
            \textsc{AmazonReview} (Polarity)& $768^{\dag}$ & $2$ & $3.6$M & $400$K & \citep{zhang2015character-level}\\
            \textsc{AmazonReview} (Full Scale) & $768^{\dag}$ & $5$ & $3$M & $650$K & \citep{zhang2015character-level}\\
            \textsc{Yahoo Answers} & $768^{\dag}$ & 10 & 1.4M & 60K & \citep{zhang2015character-level} \\
			\bottomrule
		\end{tabular}%
    }
    \caption{Summary of all the datasets used in this work. $\ast$: we rescale the \usps digits to $28\times 28$ for comparison to the *NIST datasets. $\ddag$: we rescale Tiny-ImageNet to $32\times 32$ for comparison to \textsc{cifar-10}. $\dag$: for text datasets, variable-length sentences are embedded to fixed-dimensional vectors using \textsc{bert}. }\label{tab:dataset_details}.
\end{table*}%

\section{Time Complexity Analysis}\label{sec:complexity_analysis}

For the analyses in this section, assume that $\cD_S$ and $\cD_T$ respectively have $n$ and $m$ labeled examples in $\R^d$ and $k_s, k_t$ classes. In addition, let $\mathcal{N}^{S}_{\mathcal{D}}(i) := \{ x \in \cX \mid (x,y=i) \in \mathcal{D}\}$ be the subset of examples in $\cD_S$ with label $i$, and define analogously $\mathcal{N}^{T}_{\mathcal{D}}(j)$. The denote the cardinalities of these subsets as $n_s^i \triangleq | \mathcal{N}_s^{(i)} |$ and analogously for $n_t^j$. 

Direct computation of the distance \eqref{eq:ottask_metric} involves two main steps: 
\begin{enumerate}[label=(\roman*)]
    \item computing pairwise pointwise distances (each requiring solution of a label-to-label OT sub-problem), and 
    \item a global OT problem between the two samples. 
\end{enumerate}

Step (ii) is identical for both the general distance $d_{\OT}$ and its Gaussian approximation counterpart $d_{\OT\mhyphen\cN}$, so we analyze it first. This is an OT problem between two discrete distributions of size $n$ and $m$,  which can be solved exactly in $O\bigl((n +m)nm \log(nm) \bigr)$ using interior point methods or Orlin's algorithm for the uncapacitated min cost flow problem \citep{peyre2019computational}. Alternatively, it can be solved $\tau$-approximately in $O(nm\log(\max\{n,m\}) \tau^{-3})$ time using the Sinkhorn algorithm \citep{altschuler2017nearlinear}. 

We next analyze step (i) individually for the two OTDD versions. Combined, they provide a proof of Theorem~\ref{thm:complexity}.

\subsection{Pointwise distance computation for $d_{\OT}$}\label{sec:complexity_general}

Consider a single pair of points, $(x,y=i) \in \cD_A$ and $(x',y'=j) \in \cD_B$. Evaluating $\|x -x'\|$ has $O(d)$ complexity, while $W(\alpha_y, \beta_{y'})$ is an $n_s^i \times n_t^j$ OT problem which itself requires computing a distance matrix (at cost $O(n_s^in_t^jd)$), and then solving the OT problem, which as discussed before, be done exactly in $O\bigl((n_s^i + n_t^j)n_s^in_t^j \log(n_s^i + n_t^j) \bigr)$ or $\tau$-approximately in $O(n_s^in_t^j\log(\max\{n_s^i, n_t^j\}) \tau^{-3})$.

For simplicity, let us denote $\mathfrak{n}_s=\max_{i}n_s^i$, and $\mathfrak{n}_t=\max_{j}n_t^j$ the size of the largest label cluster in each dataset, and $\mathfrak{n}=\max\{\mathfrak{n}_s,\mathfrak{n}_t\}$ the overall largest one. Using these, and combining all of the above, the overall worst case complexity for the computation of the $n\times m$ pairwise distances can be expressed as
\begin{equation}\label{eq:naive_complexity_distance}
    O\bigl(nm (d + \mathfrak{n}^3\log \mathfrak{n} + d\mathfrak{n}^2)\bigr),
\end{equation}
which is what we wanted to show. 

\qed

\subsection{Pointwise distance computation for $d_{\OT\mhyphen\cN}$}

As before, consider a pair of points $(x,y=i) \in \cD_A$ and $(x',y'=j) \in \cD_B$ whose cluster sizes are $n_s^i$ and $n_t^j$ respectively.
As mentioned in Section~\ref{sec:computation}, for $d_{\OT\mhyphen\cN}$ we first compute all the per-class means and covariance matrices. This step is clearly dominated by latter, which is $O(d^2 n_s^i)$.\footnote{technically, this would be $O(d^{\omega}n_s^i)$ where $\omega$ is the coefficient of matrix multiplication, but we take $\omega=3$ for simplicity.} Considering all labels from both datasets, this amounts to a worst-case complexity of $O\bigl(d^2(k_s \mathfrak{n}_s + k_t \mathfrak{n}_t) \bigr)$. 

Once the means and covariances have been computed, we precompute all the $k_s \times k_t$ pair-wise label-to-label distances $\W_2(\alpha_y, \beta_{y'})$ using Eq.~\eqref{eq:gaussian_wasserstein}. This computation is dominated by the matrix square roots. If done exactly, these involve a full eigendecomposition, at cost $O(d^3)$, so the total cost for this step is $O(k_sk_td^3)$. 

Finally, while computing the pairwise distance, we will incur in $O(nmd)$ to obtain $\|x=x'\|$. Putting all of these together, and replacing $\mathfrak{n}_s, \mathfrak{n}_t$ by $\mathfrak{n}$, we obtain a total cost for precomputing all the point-wise distances of:
\[ O(nmd + k_sk_td^3 + d^2\mathfrak{n}(k_s + k_t), \]
which concludes the proof. 

\qed

\section{Dataset Details}\label{sec:datasets}

Information about all the datasets used, including references, are provided in Table~\ref{tab:dataset_details}.

\section{Optimization and Training Details}\label{sec:training_details}

For the adaptation experiments on the *NIST datasets, we use a LeNet-5 architecture with ReLU nonlinearities trained for 20 epochs using ADAM with learning rate \SI{1e-3} and weight decay \SI{1e-6}. It was fine-tuned for 10 epochs on the target domain(s) using the same optimization parameters. 

For the Tiny-ImageNet to \textsc{Cifar-10} adaptation results, we use a ResNet-50 trained for 300 epochs using SGD with learning rate 0.1 momentum 0.9 and weight decay \SI{1e-4}. It was fine-tuned for 30 epochs on the target domain using SGD with same parameters except 0.01 learning rate. We discard pairs for which the variance on adaptation accuracy is beyond a certain threshold. 

For the text classification experiments, we use a pretrained \textsc{bert} architecture (the \texttt{bert-base-uncased} model of the \texttt{transformers}\footnote{\url{huggingface.co/transformers/}} library). We first embed all sentences using this model. Then, for each pair of source/target domains, we first fine-tune using ADAM with learning rate \SI{2e-5} for 10 epochs on the full source domain data, and the fine-tune on the restricted target domain data with the same optimization parameters for 2 epochs. 

Our implementation of the OTDD relies on the \texttt{pot}\footnote{\url{pot.readthedocs.io/en/stable/}} and \texttt{geomloss}\footnote{\url{www.kernel-operations.io/geomloss/}} python packages.

\section{Robustness of the Distance}\label{sec:robustness}

\vspace{1cm}
\noindent\begin{minipage}{\textwidth}
    \centering
    \includegraphics[width=0.9\linewidth, trim={0 0.5cm 0 0.5cm}, clip]{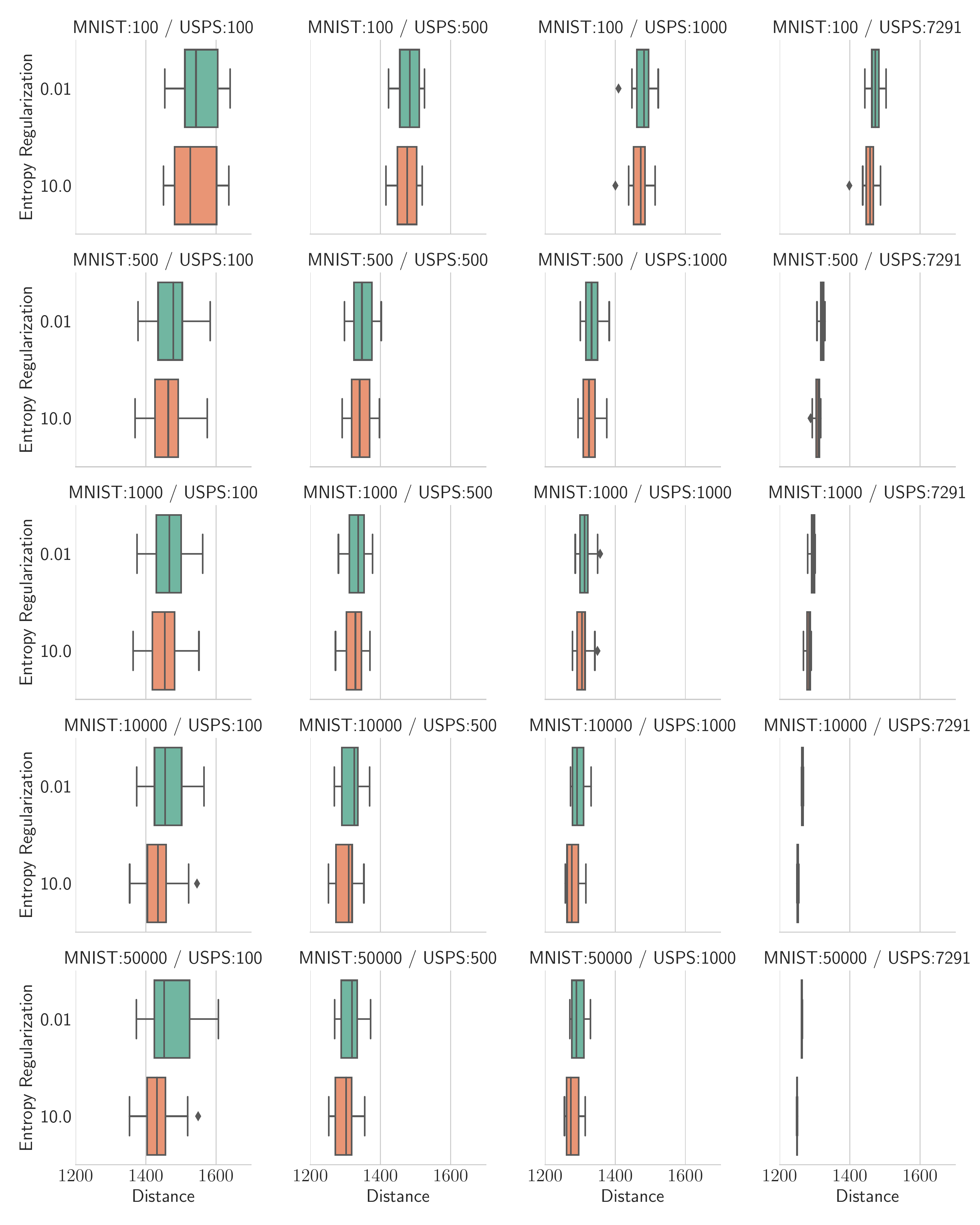}
    \captionof{figure}{\textbf{Robustness Analysis}: distances computed on subsets of varying size (rows: \mnist, columns: \usps), over 10 random repetitions, for two values of the regularization parameter $\varepsilon$.}\label{fig:mnist_usps_robustness}    
\end{minipage}

%% file: main.bbl
\begin{thebibliography}{51}
\providecommand{\natexlab}[1]{#1}
\providecommand{\url}[1]{\texttt{#1}}
\expandafter\ifx\csname urlstyle\endcsname\relax
  \providecommand{\doi}[1]{doi: #1}\else
  \providecommand{\doi}{doi: \begingroup \urlstyle{rm}\Url}\fi

\bibitem[Achille et~al.(2018)Achille, Mbeng, and Soatto]{achille2018dynamics}
Achille, A., Mbeng, G., and Soatto, S.
\newblock Dynamics and reachability of learning tasks.
\newblock October 2018.

\bibitem[Achille et~al.(2019)Achille, Lam, Tewari, Ravichandran, Maji, Fowlkes,
  Soatto, and Perona]{achille2019task}
Achille, A., Lam, M., Tewari, R., Ravichandran, A., Maji, S., Fowlkes, C.,
  Soatto, S., and Perona, P.
\newblock {Task2Vec}: Task embedding for {Meta-Learning}.
\newblock In \emph{Proceedings of the {IEEE} International Conference on
  Computer Vision}, pp.\  6430--6439, 2019.

\bibitem[Altschuler et~al.(2017)Altschuler, Niles-Weed, and
  Rigollet]{altschuler2017nearlinear}
Altschuler, J., Niles-Weed, J., and Rigollet, P.
\newblock Near-linear time approximation algorithms for optimal transport via
  sinkhorn iteration.
\newblock In Guyon, I., Luxburg, U.~V., Bengio, S., Wallach, H., Fergus, R.,
  Vishwanathan, S., and Garnett, R. (eds.), \emph{Advances in Neural
  Information Processing Systems 30}, pp.\  1964--1974. Curran Associates,
  Inc., 2017.

\bibitem[Alvarez-Melis \& Jaakkola(2018)Alvarez-Melis and
  Jaakkola]{alvarez-melis2018gromov}
Alvarez-Melis, D. and Jaakkola, T.
\newblock {Gromov-Wasserstein} alignment of word embedding spaces.
\newblock In \emph{Proceedings of the 2018 Conference on Empirical Methods in
  Natural Language Processing}, pp.\  1881--1890, 2018.
\newblock \doi{10.18653/v1/D18-1214}.

\bibitem[Alvarez-Melis et~al.(2018)Alvarez-Melis, Jaakkola, and
  Jegelka]{alvarez-melis2018structured}
Alvarez-Melis, D., Jaakkola, T.~S., and Jegelka, S.
\newblock Structured optimal transport.
\newblock In {Amos Storkey And} (ed.), \emph{Proceedings of the {Twenty-First}
  International Conference on Artificial Intelligence and Statistics},
  volume~84 of \emph{Proceedings of Machine Learning Research}, pp.\
  1771--1780. PMLR, 2018.

\bibitem[Amari(1985)]{amari1985differential}
Amari, S.-I.
\newblock \emph{{Differential-Geometrical} Methods in Statistics}, volume~28 of
  \emph{Lecture Notes in Statistics}.
\newblock Springer New York, New York, NY, 1985.
\newblock ISBN 9780387960562, 9781461250562.
\newblock \doi{10.1007/978-1-4612-5056-2}.

\bibitem[Amari(1998)]{amari1998natural}
Amari, S.-I.
\newblock Natural gradient works efficiently in learning.
\newblock \emph{Neural Comput.}, 10\penalty0 (2):\penalty0 251--276, February
  1998.
\newblock ISSN 0899-7667.
\newblock \doi{10.1162/089976698300017746}.

\bibitem[Amari \& Nagaoka(2000)Amari and Nagaoka]{amari2000methods}
Amari, S.-I. and Nagaoka, H.
\newblock \emph{Methods of Information Geometry}.
\newblock Translations of Mathematical Monographs. American Mathematical
  Society, 2000.
\newblock ISBN 9780821843024.

\bibitem[Ben-David et~al.(2007)Ben-David, Blitzer, Crammer, and
  Pereira]{ben-david2007analysis}
Ben-David, S., Blitzer, J., Crammer, K., and Pereira, F.
\newblock Analysis of representations for domain adaptation.
\newblock In Sch{\"o}lkopf, B., Platt, J.~C., and Hoffman, T. (eds.),
  \emph{Advances in Neural Information Processing Systems 19}, pp.\  137--144.
  MIT Press, 2007.

\bibitem[Chan et~al.(1983)Chan, Golub, and Leveque]{chan1983algorithms}
Chan, T.~F., Golub, G.~H., and Leveque, R.~J.
\newblock Algorithms for computing the sample variance: Analysis and
  recommendations.
\newblock \emph{Am. Stat.}, 37\penalty0 (3):\penalty0 242--247, August 1983.
\newblock ISSN 0003-1305.
\newblock \doi{10.1080/00031305.1983.10483115}.

\bibitem[Clanuwat et~al.(2018)Clanuwat, Bober-Irizar, Kitamoto, Lamb, Yamamoto,
  and Ha]{clanuwat2018deep}
Clanuwat, T., Bober-Irizar, M., Kitamoto, A., Lamb, A., Yamamoto, K., and Ha,
  D.
\newblock Deep learning for classical japanese literature.
\newblock December 2018.

\bibitem[Cohen et~al.(2017)Cohen, Afshar, Tapson, and van
  Schaik]{cohen2017emnist}
Cohen, G., Afshar, S., Tapson, J., and van Schaik, A.
\newblock {EMNIST}: Extending {MNIST} to handwritten letters.
\newblock In \emph{2017 International Joint Conference on Neural Networks
  ({IJCNN})}, pp.\  2921--2926. IEEE, May 2017.
\newblock \doi{10.1109/IJCNN.2017.7966217}.

\bibitem[Cortes \& Mohri(2011)Cortes and Mohri]{cortes2011domain}
Cortes, C. and Mohri, M.
\newblock Domain adaptation in regression.
\newblock In \emph{Algorithmic Learning Theory}, pp.\  308--323. Springer
  Berlin Heidelberg, 2011.
\newblock \doi{10.1007/978-3-642-24412-4\_25}.

\bibitem[Courty et~al.(2017)Courty, Flamary, Tuia, and
  Rakotomamonjy]{courty2017optimal}
Courty, N., Flamary, R., Tuia, D., and Rakotomamonjy, A.
\newblock Optimal transport for domain adaptation.
\newblock \emph{IEEE Trans. Pattern Anal. Mach. Intell.}, 39\penalty0
  (9):\penalty0 1853--1865, September 2017.
\newblock ISSN 0162-8828.
\newblock \doi{10.1109/TPAMI.2016.2615921}.

\bibitem[Cuturi(2013)]{cuturi2013sinkhorn}
Cuturi, M.
\newblock Sinkhorn distances: Lightspeed computation of optimal transport.
\newblock In Burges, C. J.~C., Bottou, L., Welling, M., Ghahramani, Z., and
  Weinberger, K.~Q. (eds.), \emph{Advances in Neural Information Processing
  Systems 26}, pp.\  2292--2300. Curran Associates, Inc., 2013.

\bibitem[De~Lathauwer(2003)]{delathauwer2003simultaneous}
De~Lathauwer, L.
\newblock Simultaneous matrix diagonalization: the overcomplete case.
\newblock In \emph{Proc. of the 4th International Symposium on {ICA} and Blind
  Signal Separation, Nara, Japan}, volume 8122, pp.\  825. kecl.ntt.co.jp,
  2003.

\bibitem[Delon \& Desolneux(2019)Delon and Desolneux]{delon2019wasserstein}
Delon, J. and Desolneux, A.
\newblock A wasserstein-type distance in the space of gaussian mixture models.
\newblock July 2019.

\bibitem[Deng et~al.(2009)Deng, Dong, Socher, Li, {Kai Li}, and {Li
  Fei-Fei}]{deng2009imagenet}
Deng, J., Dong, W., Socher, R., Li, L., {Kai Li}, and {Li Fei-Fei}.
\newblock {ImageNet}: A large-scale hierarchical image database.
\newblock In \emph{2009 {IEEE} Conference on Computer Vision and Pattern
  Recognition}, pp.\  248--255. IEEE, June 2009.
\newblock \doi{10.1109/CVPR.2009.5206848}.

\bibitem[Devlin et~al.(2019)Devlin, Chang, Lee, and Toutanova]{devlin2019bert}
Devlin, J., Chang, M.-W., Lee, K., and Toutanova, K.
\newblock {BERT}: Pre-training of deep bidirectional transformers for language
  understanding.
\newblock In \emph{Proceedings of the 2019 Conference of the North American
  Chapter of the Association for Computational Linguistics: Human Language
  Technologies, Volume 1 (Long and Short Papers)}, pp.\  4171--4186, 2019.

\bibitem[Dukler et~al.(2019)Dukler, Li, Lin, and
  Montufar]{dukler2019wasserstein}
Dukler, Y., Li, W., Lin, A., and Montufar, G.
\newblock {W}asserstein of {W}asserstein loss for learning generative models.
\newblock In Chaudhuri, K. and Salakhutdinov, R. (eds.), \emph{Proceedings of
  the 36th International Conference on Machine Learning}, volume~97 of
  \emph{Proceedings of Machine Learning Research}, pp.\  1716--1725, Long
  Beach, California, USA, 2019. PMLR.

\bibitem[Feydy et~al.(2019)Feydy, S{\'e}journ{\'e}, Vialard, Amari, Trouve, and
  Peyr{\'e}]{feydy2019interpolating}
Feydy, J., S{\'e}journ{\'e}, T., Vialard, F.-X., Amari, S.-I., Trouve, A., and
  Peyr{\'e}, G.
\newblock Interpolating between optimal transport and {MMD} using sinkhorn
  divergences.
\newblock In Chaudhuri, K. and Sugiyama, M. (eds.), \emph{Proceedings of the
  22nd International Conference on Artificial Intelligence and Statistics},
  volume~89 of \emph{Proceedings of Machine Learning Research}, pp.\
  2681--2690. PMLR, 2019.

\bibitem[Frogner et~al.(2019)Frogner, Mirzazadeh, and
  Solomon]{frogner2019learning}
Frogner, C., Mirzazadeh, F., and Solomon, J.
\newblock Learning embeddings into entropic wasserstein spaces.
\newblock In \emph{International Conference on Learning Representations}, May
  2019.

\bibitem[Gelbrich(1990)]{gelbrich1990formula}
Gelbrich, M.
\newblock On a formula for the {L2} wasserstein metric between measures on
  euclidean and hilbert spaces.
\newblock \emph{Math. Nachr.}, 147\penalty0 (1):\penalty0 185--203, November
  1990.
\newblock ISSN 0025-584X.
\newblock \doi{10.1002/mana.19901470121}.

\bibitem[Genevay et~al.(2018)Genevay, Peyre, and Cuturi]{genevay2018learning}
Genevay, A., Peyre, G., and Cuturi, M.
\newblock Learning generative models with sinkhorn divergences.
\newblock In Storkey, A. and Perez-Cruz, F. (eds.), \emph{Proceedings of the
  {Twenty-First} International Conference on Artificial Intelligence and
  Statistics}, volume~84 of \emph{Proceedings of Machine Learning Research},
  pp.\  1608--1617, Playa Blanca, Lanzarote, Canary Islands, 2018. PMLR.

\bibitem[Genevay et~al.(2019)Genevay, Chizat, Bach, Cuturi, and
  Peyr{\'e}]{genevay2019sample}
Genevay, A., Chizat, L., Bach, F., Cuturi, M., and Peyr{\'e}, G.
\newblock Sample complexity of sinkhorn divergences.
\newblock In Chaudhuri, K. and Sugiyama, M. (eds.), \emph{Proceedings of
  Machine Learning Research}, volume~89 of \emph{Proceedings of Machine
  Learning Research}, pp.\  1574--1583. PMLR, 2019.

\bibitem[Higham(2008)]{higham2008functions}
Higham, N.~J.
\newblock \emph{Functions of Matrices: Theory and Computation}.
\newblock SIAM, January 2008.
\newblock ISBN 9780898717778.

\bibitem[Hull(1994)]{hull1994database}
Hull, J.~J.
\newblock A database for handwritten text recognition research.
\newblock \emph{IEEE Trans. Pattern Anal. Mach. Intell.}, 16\penalty0
  (5):\penalty0 550--554, May 1994.
\newblock ISSN 0162-8828, 1939-3539.
\newblock \doi{10.1109/34.291440}.

\bibitem[Kantorovitch(1942)]{kantorovitch1942translocation}
Kantorovitch, L.
\newblock On the translocation of masses.
\newblock \emph{Dokl. Akad. Nauk SSSR}, 37\penalty0 (7-8):\penalty0 227--229,
  1942.
\newblock ISSN 0002-3264.

\bibitem[Khodak et~al.(2019)Khodak, Balcan, and Talwalkar]{khodak2019adaptive}
Khodak, M., Balcan, M.-F.~F., and Talwalkar, A.~S.
\newblock Adaptive {Gradient-Based} {Meta-Learning} methods.
\newblock In Wallach, H., Larochelle, H., Beygelzimer, A., d'Alch{\'e} Buc, F.,
  Fox, E., and Garnett, R. (eds.), \emph{Advances in Neural Information
  Processing Systems 32}, pp.\  5915--5926. Curran Associates, Inc., 2019.

\bibitem[Krizhevsky \& Hinton(2009)Krizhevsky and
  Hinton]{krizhevsky2009learning}
Krizhevsky, A. and Hinton, G.
\newblock Learning multiple layers of features from tiny images.
\newblock 2009.

\bibitem[Kuhn et~al.(2019)Kuhn, Esfahani, Nguyen, and
  Shafieezadeh-Abadeh]{kuhn2019wasserstein}
Kuhn, D., Esfahani, P.~M., Nguyen, V.~A., and Shafieezadeh-Abadeh, S.
\newblock Wasserstein distributionally robust optimization: Theory and
  applications in machine learning.
\newblock August 2019.

\bibitem[Le et~al.(2019)Le, Yamada, Fukumizu, and Cuturi]{le2019tree-sliced}
Le, T., Yamada, M., Fukumizu, K., and Cuturi, M.
\newblock {Tree-Sliced} variants of wasserstein distances.
\newblock In Wallach, H., Larochelle, H., Beygelzimer, A., {\textbackslash'e},
  Fox, E., and Garnett, R. (eds.), \emph{Advances in Neural Information
  Processing Systems 32}, pp.\  12283--12294. Curran Associates, Inc., 2019.

\bibitem[LeCun et~al.(2010)LeCun, Cortes, and Burges]{lecun2010mnist}
LeCun, Y., Cortes, C., and Burges, C.~J.
\newblock {MNIST} handwritten digit database.
\newblock 2010.

\bibitem[Leite \& Brazdil(2005)Leite and Brazdil]{leite2005predicting}
Leite, R. and Brazdil, P.
\newblock Predicting relative performance of classifiers from samples.
\newblock In \emph{Proceedings of the 22nd international conference on Machine
  learning}, pp.\  497--503. dl.acm.org, 2005.

\bibitem[Li(2006)]{li2006data}
Li, L.
\newblock \emph{Data Complexity in Machine Learning and Novel Classification
  Algorithms}.
\newblock PhD thesis, California Institute of Technology, 2006.

\bibitem[Liang et~al.(2019)Liang, Poggio, Rakhlin, and
  Stokes]{liang2019fisher-rao}
Liang, T., Poggio, T., Rakhlin, A., and Stokes, J.
\newblock {Fisher-Rao} metric, geometry, and complexity of neural networks.
\newblock In \emph{Proceedings of the 22nd International Conference on
  Artificial Intelligence and Statistics}. PMLR, 2019.

\bibitem[Mallasto \& Feragen(2017)Mallasto and Feragen]{mallasto2017learning}
Mallasto, A. and Feragen, A.
\newblock Learning from uncertain curves: The 2-wasserstein metric for gaussian
  processes.
\newblock In Guyon, I., Luxburg, U.~V., Bengio, S., Wallach, H., Fergus, R.,
  Vishwanathan, S., and Garnett, R. (eds.), \emph{Advances in Neural
  Information Processing Systems 30}, pp.\  5660--5670. Curran Associates,
  Inc., 2017.

\bibitem[Mansour et~al.(2009)Mansour, Mohri, and
  Rostamizadeh]{mansour2009domain}
Mansour, Y., Mohri, M., and Rostamizadeh, A.
\newblock Domain adaptation: Learning bounds and algorithms.
\newblock In \emph{The 22nd Conference on Learning Theory}. arxiv.org, 2009.

\bibitem[M{\'e}moli(2011)]{memoli2011gromov}
M{\'e}moli, F.
\newblock {Gromov--Wasserstein} distances and the metric approach to object
  matching.
\newblock \emph{Found. Comput. Math.}, 11\penalty0 (4):\penalty0 417--487,
  August 2011.
\newblock ISSN 1615-3375, 1615-3383.
\newblock \doi{10.1007/s10208-011-9093-5}.

\bibitem[M{\'e}moli(2017)]{memoli2017distances}
M{\'e}moli, F.
\newblock Distances between datasets.
\newblock In Najman, L. and Romon, P. (eds.), \emph{Modern Approaches to
  Discrete Curvature}, pp.\  115--132. Springer International Publishing, Cham,
  2017.
\newblock ISBN 9783319580029.
\newblock \doi{10.1007/978-3-319-58002-9\_3}.

\bibitem[Muzellec \& Cuturi(2018)Muzellec and Cuturi]{muzellec2018generalizing}
Muzellec, B. and Cuturi, M.
\newblock Generalizing point embeddings using the wasserstein space of
  elliptical distributions.
\newblock In Bengio, S., Wallach, H., Larochelle, H., Grauman, K.,
  Cesa-Bianchi, N., and Garnett, R. (eds.), \emph{Advances in Neural
  Information Processing Systems 31}, pp.\  10237--10248. Curran Associates,
  Inc., 2018.

\bibitem[Peters et~al.(2018)Peters, Neumann, Iyyer, Gardner, Clark, Lee, and
  Zettlemoyer]{peters2018deep}
Peters, M., Neumann, M., Iyyer, M., Gardner, M., Clark, C., Lee, K., and
  Zettlemoyer, L.
\newblock Deep contextualized word representations.
\newblock In \emph{Proceedings of the 2018 Conference of the North American
  Chapter of the Association for Computational Linguistics: Human Language
  Technologies, Volume 1 (Long Papers)}, pp.\  2227--2237. aclweb.org, 2018.

\bibitem[Peyr{\'e} \& Cuturi(2019)Peyr{\'e} and Cuturi]{peyre2019computational}
Peyr{\'e}, G. and Cuturi, M.
\newblock Computational optimal transport.
\newblock \emph{Foundations and Trends\textregistered{} in Machine Learning},
  11\penalty0 (5-6):\penalty0 355--607, 2019.
\newblock ISSN 1935-8237.
\newblock \doi{10.1561/2200000073}.

\bibitem[Radford et~al.(2019)Radford, Wu, Amodei, Amodei, Clark, Brundage, and
  Sutskever]{radford2019better}
Radford, A., Wu, J., Amodei, D., Amodei, D., Clark, J., Brundage, M., and
  Sutskever, I.
\newblock Better language models and their implications.
\newblock \emph{OpenAI Blog https://openai. com/blog/better-language-models},
  2019.

\bibitem[Rubner et~al.(2000)Rubner, Tomasi, and Guibas]{rubner2000earth}
Rubner, Y., Tomasi, C., and Guibas, L.~J.
\newblock The earth mover's distance as a metric for image retrieval.
\newblock \emph{Int. J. Comput. Vis.}, 40\penalty0 (2):\penalty0 99--121,
  November 2000.
\newblock ISSN 0920-5691, 1573-1405.
\newblock \doi{10.1023/A:1026543900054}.

\bibitem[Tran et~al.(2019)Tran, Nguyen, and Hassner]{tran2019transferability}
Tran, A.~T., Nguyen, C.~V., and Hassner, T.
\newblock Transferability and hardness of supervised classification tasks.
\newblock August 2019.

\bibitem[Villani(2003)]{villani2003topics}
Villani, C.
\newblock \emph{Topics in Optimal Transportation}.
\newblock American Mathematical Soc., 2003.
\newblock ISBN 9780821833124.

\bibitem[Villani(2008)]{villani2008optimal}
Villani, C.
\newblock \emph{Optimal transport, Old and New}, volume 338.
\newblock Springer Science \& Business Media, 2008.
\newblock ISBN 9783540710493.

\bibitem[Xiao et~al.(2017)Xiao, Rasul, and Vollgraf]{xiao2017fashion-mnist}
Xiao, H., Rasul, K., and Vollgraf, R.
\newblock {Fashion-MNIST}: a novel image dataset for benchmarking machine
  learning algorithms.
\newblock August 2017.

\bibitem[Yurochkin et~al.(2019)Yurochkin, Claici, Chien, Mirzazadeh, and
  Solomon]{yurochkin2019hierarchical}
Yurochkin, M., Claici, S., Chien, E., Mirzazadeh, F., and Solomon, J.~M.
\newblock Hierarchical optimal transport for document representation.
\newblock In Wallach, H., Larochelle, H., Beygelzimer, A., d'Alch{\'e} Buc, F.,
  Fox, E., and Garnett, R. (eds.), \emph{Advances in Neural Information
  Processing Systems 32}, pp.\  1599--1609. Curran Associates, Inc., 2019.

\bibitem[Zhang et~al.(2015)Zhang, Zhao, and LeCun]{zhang2015character-level}
Zhang, X., Zhao, J., and LeCun, Y.
\newblock Character-level convolutional networks for text classification.
\newblock In Cortes, C., Lawrence, N.~D., Lee, D.~D., Sugiyama, M., and
  Garnett, R. (eds.), \emph{Advances in Neural Information Processing Systems
  28}, pp.\  649--657. Curran Associates, Inc., 2015.

\end{thebibliography}
